\documentclass{article}
\usepackage{enumitem}
\usepackage[preprint]{corl_2026} % Uncomment for pre-prints (e.g., arxiv); This is like ``final'', but will remove the CORL footnote.

\usepackage{amssymb,amsmath,amsfonts,bm}
\usepackage{graphicx}
\graphicspath{{./}{../}}
\usepackage[table]{xcolor}  % 1
\usepackage{hyperref}       % 2
\usepackage{cleveref}       % 3
\usepackage{tabularx}
\usepackage{array}
\usepackage{booktabs}
\usepackage[table]{xcolor}
\definecolor{darkblue}{rgb}{0.0, 0.0, 0.5}
\hypersetup{
  colorlinks=true,
  linkcolor=darkblue,
  citecolor=darkblue,
  urlcolor=darkblue
}
\usepackage{mathtools}
\usepackage{wrapfig}
\usepackage{algorithmic}
\usepackage{algorithm}
\usepackage{booktabs}
\usepackage{tikz}
\usepackage{multirow}
\usepackage[dvipsnames]{xcolor}
\usetikzlibrary{arrows.meta,positioning,calc,fit,shapes.geometric,backgrounds,decorations.pathreplacing}

\usepackage{tabularx}
\usepackage{enumitem}
\usepackage[most]{tcolorbox}
\newlist{tabenum}{enumerate}{1}
\setlist[tabenum]{label=(\arabic*), leftmargin=*, nosep, topsep=0pt, partopsep=0pt}

\title{HARBOR: A Harness Framework for Agentic Robot Reinforcement Learning}

\author{
  Zechu Li~$^{1,7}$ ~~~~Yufeng Jin~$^{1,2}$ ~~~~Xiaoyang Liu~$^3$ ~~~~\textbf{Puze Liu}~$^{4,5}$\thanks{Corresponding author} ~~~~\textbf{Vignesh Prasad}~$^1$\\
   \textbf{Carlo D'Eramo}~$^{6,7}$ 
  ~~~~ \textbf{Georgia Chalvatzaki}~$^{1,7}$\\
  TU Darmstadt~$^1$~~~~Honda Research Institute Europe~$^2$
  ~~~~Columbia University~$^3$
  \\
  Tongji University~$^4$~~~~Shanghai Research Institute for Intelligent Autonomous Systems~$^5$
  \\
  University of Würzburg~$^6$
  ~~~~Hessian.AI~$^7$
  \vspace{-5mm}
}

\begin{document}
\maketitle

%===============================================================================

\begin{abstract}
Reinforcement learning (RL) has become a powerful paradigm for robot learning, particularly in sim-to-real settings, but its broader adoption remains limited by the engineering pipeline surrounding the algorithms. Building tasks, shaping rewards, and tuning hyperparameters require substantial expert effort, making RL workflows costly and difficult to scale. We introduce \textit{HARBOR}, an agentic framework that frames robot RL automation as a harness-engineering problem: given a simulator codebase and a task specification, it automates the workflow from environment setup to policy training in simulation. HARBOR decomposes such high-level objectives into bounded stages executed by specialized agents through standardized commands, persistent artifacts, executable gates, and reusable knowledge, and scales iteration via decentralized parallel trials and experience learning across runs. We evaluate HARBOR across 6 benchmarks and 16 tasks in total, spanning manipulation, locomotion, and bimanual dexterous control. We demonstrate that HARBOR automates the simulation RL workflow end-to-end, designs rewards, tunes algorithms to match or improve over default configurations, and reduces engineering effort at practical token and wall-clock cost; the resulting policies can also be transferred to real robots.
\end{abstract}
\vspace{-2mm}

%===============================================================================

\begin{figure}[t!]
    \centering
    \includegraphics[width=\textwidth]{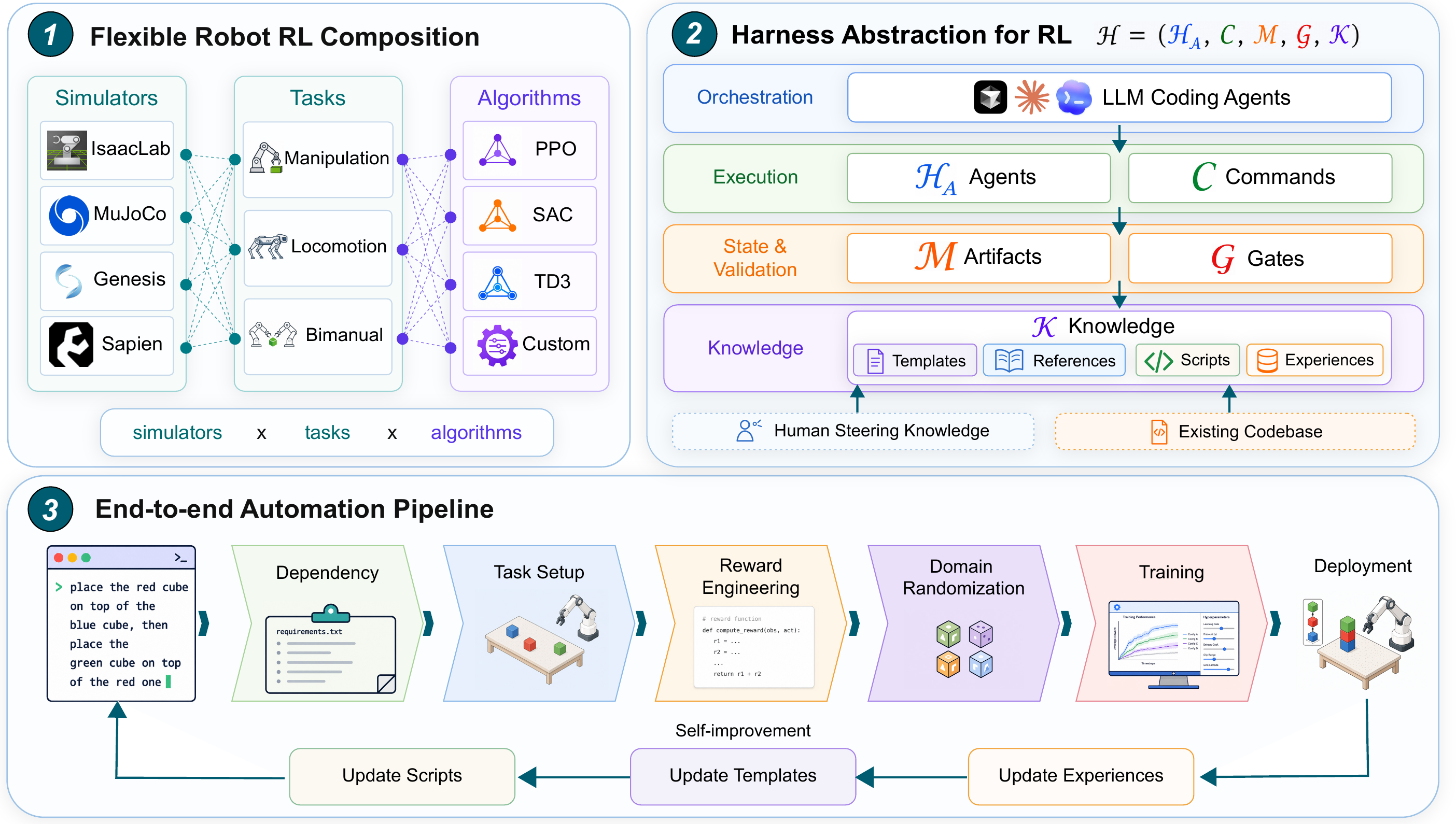}
    \vspace{-3mm}
    \caption{
    Overview of HARBOR. (1) HARBOR supports arbitrary simulator × task × algorithm combinations across diverse benchmarks. (2) HARBOR treats robot RL automation as an RL harness that coordinates specialized agents, executable commands, mutable artifacts,
    verifiable gates, and reusable knowledge. (3) A natural-language request flows from dependency setup to training, with a self-improvement loop updating reusable knowledge across runs.
    }
    \label{fig:teaser}
    \vspace{-3mm}
\end{figure}

\section{Introduction}
Reinforcement learning (RL)~\citep{sutton1998reinforcement, kaelbling1996reinforcement} has become a dominant paradigm for robot learning, particularly in sim-to-real settings where policies are trained in simulation and transferred to physical robots~\citep{zhao2020sim, torne2024reconciling}. The standard recipe is well established: build a digital twin, design rewards, randomize dynamics, train and deploy. While this pipeline has enabled impressive results from legged locomotion~\citep{margolis2024rapid, li2025reinforcement} to dexterous manipulation~\citep{chen2023visual, li2025morphologically, lin2024twisting}, its broader adoption is increasingly limited not by the RL algorithms themselves, but by the engineering pipeline around them. Practitioners must manually construct tasks, shape rewards, calibrate domain randomization, and tune hyperparameters, often spending days to weeks on a single new setup. These specialized, labor-intensive steps make RL workflows costly and difficult to scale, limiting their effectiveness for both rapid single-task deployment and large-scale generalist robot learning.

Automating individual steps of this pipeline has been a productive line of research. Prior work has explored language-model-guided reward synthesis~\citep{ma2024eureka, hazra2025revolve}, automated domain randomization~\citep{ma2024dreureka, zeng2026dexsim2real}, and AutoRL methods for tuning algorithmic hyperparameters~\citep{liaw2018tune, akiba2019optuna, parker2022automated, afshar2022automated}. However, these systems typically target isolated stages rather than the full RL workflow. Practitioners must still configure the simulator, connect generated rewards to training code, launch and monitor sweeps on compute clusters, inspect rollouts, and transfer successful design choices across tasks by hand. As a result, each automation module reduces effort locally, but the surrounding integration burden is repeatedly paid across simulators, tasks, and algorithms.

Large language model (LLM) agents offer a promising substrate for automating the RL engineering pipeline, but long-horizon autonomy requires more than a capable model. Failures often stem from underspecified executions: agents lack the tools, abstractions, and feedback needed to reliably pursue high-level goals~\citep{meng2026agent, he2026harness, zhou2026externalization}. \emph{Harness engineering} addresses this by shifting human effort from manually executing each step to designing structured, agent-readable workflows with verifiable interfaces~\citep{lee2026meta, jin2026nautilus}. RL for robotics (robot RL) is particularly well suited to this view: the Markov Decision Process (MDP)~\citep{sutton1998reinforcement} exposes stable interfaces, namely, state, action, reward, dynamics, and termination, while rollout behaviors, reward signals, and training curves provide executable feedback, making robot RL automation a natural fit for harness engineering.

To this end, we introduce \textbf{HARBOR}, a \underline{h}arness framework for \underline{a}gentic \underline{r}o\underline{bo}t \underline{r}einforcement learning. Given a simulator codebase and a task specification, a centralized main agent decomposes the request into bounded stages, from dependency setup to policy training, and dispatches specialized agents to each. The agents invoke standardized commands grounded in built-in templates, scripts, and prior experience, and advance only after executable gates verify outputs, preventing errors from propagating downstream. Iterative stages, such as reward and algorithm tuning, run as parallel trials whose outcomes are distilled into reusable experience that later agents can retrieve to improve efficiency and reliability. We evaluate HARBOR across 6 benchmarks and 16 tasks spanning manipulation, locomotion, and bimanual dexterous control, showing that it automates the simulation RL workflow, matches or improves over default configurations through autonomous tuning, and reduces engineering effort at practical cost, with the resulting policies transferring to real robots.

Our contributions are threefold. First, we frame robot RL automation as a harness-engineering problem and identify the RL-specific interfaces and feedback signals that make long-horizon automation feasible. Second, we instantiate this view in HARBOR, a system that turns natural-language robot-learning objectives into automated and reliable RL workflows through agents, commands, artifacts, gates, and knowledge. Third, we implement HARBOR as a documented and accessible LLM-agent plugin that lowers the barrier to robot RL: its inspectable artifacts and gated checkpoints let practitioners audit, step in, and adapt the pipeline beyond expert use. We defer a full discussion of related work to Appx.~\ref{app:related} and background to Appx.~\ref{app:background}.

%===============================================================================

\section{Robot RL Automation as a Harness Engineering Problem}
\label{sec:rl_harness}

Robot RL automation is not simply a code-generation problem, but the challenge of providing a reliable workflow that can make decisions, preserve state, and verify progress over a long horizon. Robot RL pipelines are tightly coupled: task design affects rewards, rewards shape learning, and deployment failures may require revisiting earlier stages. This coupling makes unconstrained agentic automation brittle, and it also exposes structure that a harness can exploit.

\paragraph{From MDPs to Robot RL Workflows} Reinforcement learning is typically formalized as a Markov decision process (MDP)~\citep{sutton1998reinforcement, kaelbling1996reinforcement}: 
\[
    \mathcal{M} = (\mathcal{S}, \mathcal{A}, P, r, \rho_0, \gamma, \mathcal{T}),
\]
where \(\mathcal{S}\) and \(\mathcal{A}\) denote the state and action spaces, \(P\) is the transition dynamics, \(r\) is the reward function, \(\rho_0\) is the initial-state distribution, \(\gamma\) is the discount factor, and \(\mathcal{T}\) specifies termination conditions. While compact, this mathematical formulation hides the engineering complexity required to instantiate an RL problem in robotics. The state and action spaces must match simulator observations and robot control interfaces, while transition dynamics are determined by assets, control frequency, and physical parameters. Beyond the MDP itself, a complete robot RL workflow also requires algorithm integration, configuration, logging, evaluation, and, in sim-to-real settings, deployment interfaces. Robot RL is therefore not only policy optimization over a fixed MDP, but also the construction, validation, and iterative refinement of the MDP and its surrounding pipeline.

\paragraph{Harnesses for Long-Horizon Agentic Automation} A harness is the structured execution environment around an LLM agent: it defines how the agent accesses tools, preserves artifacts, observes feedback, and verifies progress~\citep{meng2026agent, lee2026meta}. This structure is crucial for long-horizon autonomy, where success depends both on model capability and on how open-ended objectives are decomposed, executed, and checked. Harness engineering therefore shifts effort from manually executing each step to designing agent-readable workflows with reusable actions and executable validations.

Robot RL automation is particularly amenable to such a harness. 
Its MDP formulation exposes relatively stable interfaces, while simulators and training runs provide executable feedback for checking intermediate progress. Interface checks can catch invalid resets, observations, and actions; diagnostic states and short training runs can expose reward or optimization failures; and rendered rollouts can reveal behavioral errors that scalar returns may hide. These checks are not complete guarantees, but provide a practical substrate for agents to validate artifacts and recover from failures.

\paragraph{Harness Abstractions for RL}
Motivated by these properties, HARBOR specializes a general agentic harness pattern to robot RL automation, as shown in Fig.~\ref{fig:teaser}(2):
\[
\mathcal{H}_{\textsc{RL}}
=
(\mathcal{H}_{A}, \mathcal{C}, \mathcal{M}, \mathcal{G}, \mathcal{K}),
\]
where \(\mathcal{H}_{A}\) denotes agents, \(\mathcal{C}\) commands, \(\mathcal{M}\) mutable artifacts, \(\mathcal{G}\) verifiable gates, and \(\mathcal{K}\) reusable knowledge. Knowledge and mutable artifacts provide context for an agent; the agent invokes commands that transform artifacts; gates validate the resulting workflow state using executable RL evidence; and passed or failed outcomes are summarized back into knowledge. In this sense, HARBOR does not guarantee semantic correctness of the final policy. Instead, it turns many common RL engineering failures into observable gate failures before they propagate downstream.
\begin{itemize}[leftmargin=*]
    \item \textbf{\texttt{Agents} \((\mathcal{H}_{A})\)} are context-isolated subprocesses assigned to bounded stages of the RL workflow. Each agent operates on stage-local artifacts and retrieved knowledge, performs the local implementation, then returns a compact summary to the main controller.

    \item \textbf{\texttt{Commands} \((\mathcal{C})\)} are reproducible operations exposed to agents, ranging from primitive calls such as \texttt{rl-sweep} to composed loops such as \texttt{tune-reward}. The same command surface can be invoked by different agents with different artifacts and gates.

    \item \textbf{\texttt{Mutable Artifacts} \((\mathcal{M})\)} externalize workflow state into persistent, inspectable objects. They serve as the communication substrate between agents and commands, reducing reliance on transient LLM context.

    \item \textbf{\texttt{Verifiable Gates} \((\mathcal{G})\)} are executable checks that determine whether a stage can advance. They include both hard interface checks and softer semantic checks, such as import and rollout.

    \item \textbf{\texttt{Reusable Knowledge} \((\mathcal{K})\)} includes templates, references, scripts, human heuristics, and accumulated experience from previous runs. It constrains generation, encodes simulator- and algorithm-specific contracts, and gives later agents access to the outcomes of earlier attempts.

\end{itemize}

% Together, these components provide an operational interface for harnessed RL development: agents propose changes, commands execute them, gates validate them, and experience accumulates as reusable knowledge.
%===============================================================================

\vspace{-3mm}
\section{HARBOR: An Agentic RL Harness Framework for Robot Learning}
\label{sec:HARBOR}

HARBOR instantiates the RL harness as an artifact-centric execution graph for robot learning. Given a user request, HARBOR decomposes the request into bounded stages. Each stage is handled by a specialized agent that invokes standardized commands, produces persistent artifacts, and advances only after executable gates validate its outputs. Tab.~\ref{tab:harbor_component} summarizes how the abstract harness tuple is instantiated as concrete workflow stages; full descriptions are provided in Appx.~\ref{app:agent_list}.

\paragraph{System Overview and Design Choices}
HARBOR supports the full lifecycle of robot RL development in simulation, including dependency setup, task construction, reward design, domain randomization, algorithm integration, and hyperparameter tuning. Users may specify any subset of the simulator, task, algorithm, training budget, or tuning objective; HARBOR infers missing choices from framework experience and codebase templates. It then materializes the workflow as runnable code, configuration files, validation logs, checkpoints, videos, metrics, and tuning summaries.

HARBOR makes three design choices that distinguish it from generic LLM coding agents. First, the workflow is artifact-centric: persistent files are the communication substrate, capturing MDP state and tuning history that persists across iterations and reducing reliance on transient LLM context. Second, gates operate at stage granularity and leverage RL's executable signals to validate each stage. Third, execution is centralized in planning but decentralized in execution, allowing parallel trials for RL tuning without main-agent context bloat and improving the time-efficiency.

\begin{table}[h]
\centering
\small
\setlength{\tabcolsep}{3pt}
\renewcommand{\arraystretch}{1}
\rowcolors{2}{gray!5}{white}
\begin{tabularx}{\linewidth}{
>{\centering\arraybackslash}p{1.95cm}
>{\raggedright\arraybackslash}X
>{\raggedright\arraybackslash}p{2.75cm}
>{\raggedright\arraybackslash}p{3.55cm}
}
\toprule
\rowcolor{white}
\textbf{Stage} & \textbf{Agent / Commands} & \textbf{Artifacts} & \textbf{Gates} \\
\midrule
Dependency setup 
& Dependency-generator; \texttt{probe-env} 
& install log, env file 
& import, device \\

Task generation 
& Task-generator; \texttt{probe-task}, \texttt{render} 
& task spec, env code 
& reset/step, obs./act. shape, render \\

Reward generation 
& Reward-generator; \texttt{tune-reward}, \texttt{rl-run}, \texttt{rl-sweep}, \texttt{rl-render}
& reward code, iteration log 
& diagnostics, rollout, video \\

RL integration 
& Integration-generator; \texttt{add-trick}, \texttt{add-log}, \texttt{rl-run} 
& train script, config 
& short train, logging, checkpoint \\

Domain randomization 
& DR-generator; \texttt{tune-dr}, \texttt{rl-run}, \texttt{rl-sweep}
& DR ranges, transfer report 
& env check, render \\

RL tuning 
& \texttt{rl-tune}, \texttt{rl-sweep}, \texttt{rl-render} 
& tuned config, curves
& rollout, cluster sample job \\
\bottomrule
\end{tabularx}
\caption{HARBOR component inventory. Each stage is implemented by an agent-command pair, recorded as artifacts, and validated by executable gates.}
\label{tab:harbor_component}
\vspace{-5mm}
\end{table}

\paragraph{Gate-Checked Execution Protocol}
HARBOR executes each stage with the same protocol. The main agent retrieves relevant knowledge and current artifacts, spawns a stage-local agent with bounded context, and asks it to edit or create artifacts through standardized commands. The command output is then evaluated by gates. If the gate passes, HARBOR commits the artifact and writes logs, metrics, videos, decisions, and failure summaries back into reusable knowledge. If a gate fails, HARBOR returns the failure summary (failed check, error message, observed values) to the stage agent for repair; after a fixed retry budget is exhausted, the stage is marked unresolved and the main agent asks the user for intervention.

For example, task-generator decomposes the MDP into ordered sub-stages, including scene construction, initialization and termination, action and observation design. Each sub-stage exposes gates before the next component is generated. For a delta end-effector pose controller, the action gate checks whether random actions produce the expected target-pose commands and the commanded target pose matches the actual pose, revealing controller, dynamics, or inverse-kinematics errors. These conservative checks do not prove semantic correctness, but they verify local interface behavior and prevent low-level task bugs from propagating into later stages.

\begin{figure}[t!]
    \centering
    \includegraphics[width=0.85\textwidth]{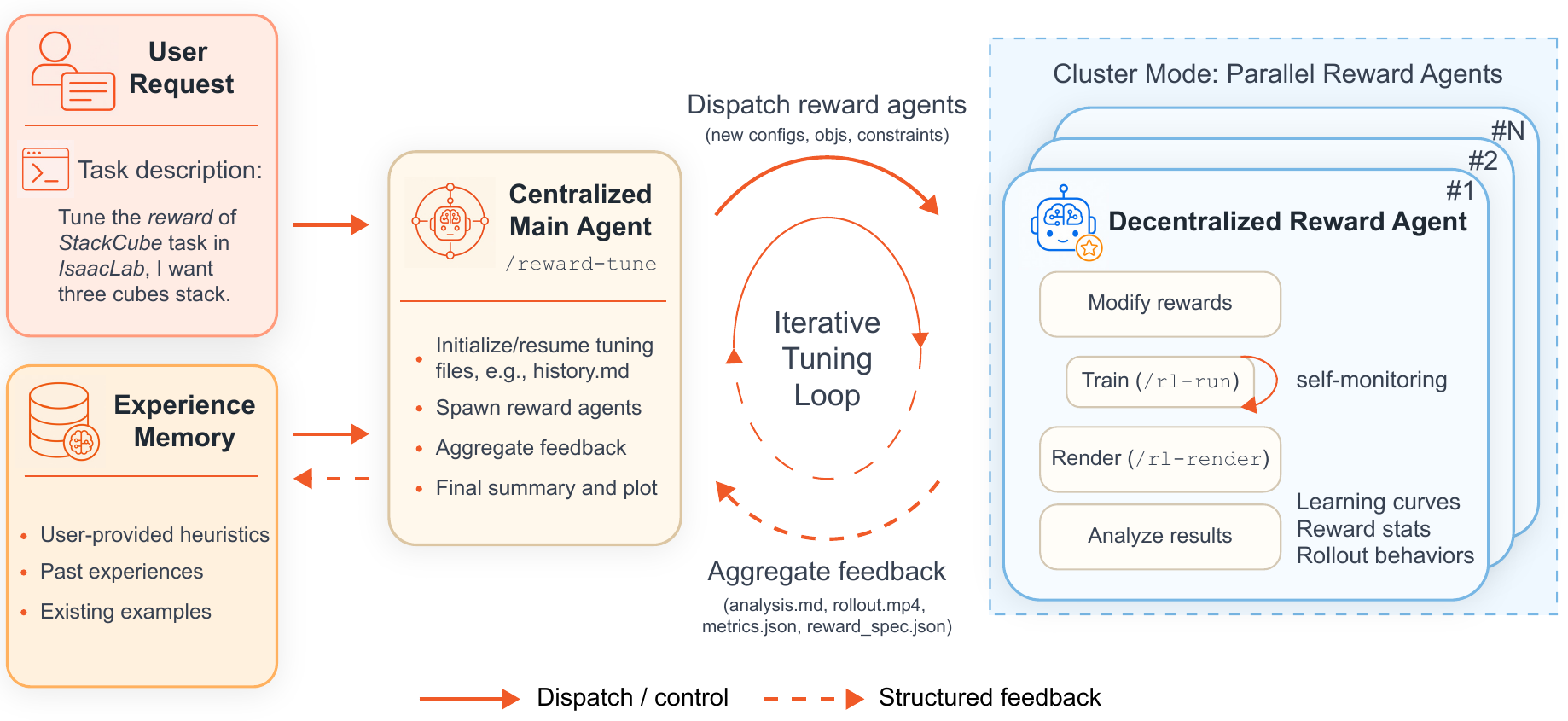}
    \vspace{-1mm}
    \caption{Centralized-control and decentralized-execution in reward tuning. Parallel agents run isolated trials and return structured feedback for aggregation, selection, and experience update.
    }
    \label{fig:tuning}
    \vspace{-3mm}
\end{figure}

\paragraph{Parallel Tuning with Experience Learning}
Reward design, domain randomization, and algorithm tuning are iterative processes that often require trying several alternatives, understanding why some trials fail, and using that evidence to decide the next change. HARBOR supports this process with centralized control and decentralized execution, as illustrated for reward design in Fig.~\ref{fig:tuning}. The main agent keeps the tuning history, which records each proposed modification and its resulting outcomes, and dispatches parallel sub-agents that each operate in isolated trial directories. Full rollout videos, reward code, and logs stay in those directories, so trials can run asynchronously, avoid overwriting shared artifacts, and be scheduled on a compute cluster. After trials finish, HARBOR aggregates and analyzes their outcomes, then decides what intervention to try next.

These traces also become the basis for experience learning. 
Rather than treating each trial as disposable context, after each tuning round, the main agent distills recurring patterns across trials into short bullet points, such as effective reward terms, unstable parameter ranges, or common failure modes. Before each new tuning round, HARBOR retrieves stage-matched experience for the current project, filtered by simulator, task, or algorithm tags. After the whole stage finishes, these points are appended back to HARBOR's experience memory, allowing later agents to reuse successful patterns and avoid known failure modes across runs.

\paragraph{Plugin Interface and Controllability} HARBOR is packaged as an LLM-agent plugin whose commands, artifacts, and gates are written in structured natural language rather than opaque code, so users can see what every stage does and what each gate checks, and edit these definitions to customize the workflow. Since every stage pauses at a gate and writes persistent artifacts, users can audit outputs, step in to correct a stage, and resume from the last gate-passing state. Users can also invoke individual agents or commands directly to build custom workflows.

% \paragraph{Implementation Details}
% HARBOR is implemented as an LLM-agent plugin that exposes the commands, artifacts, and gates described above. 
% Users interact through natural-language requests, while the workflow state remains inspectable through generated files and command histories. 
% This keeps automation controllable: users can intervene when needed and resume from the last gate-passing state. 
% Because commands, artifacts, and validation results are preserved, completed runs can be reproduced or adapted to new tasks.

%===============================================================================

\section{Experiments}
\label{sec:result}
\begin{figure}[t!]
    \centering
    \includegraphics[width=0.9\textwidth]{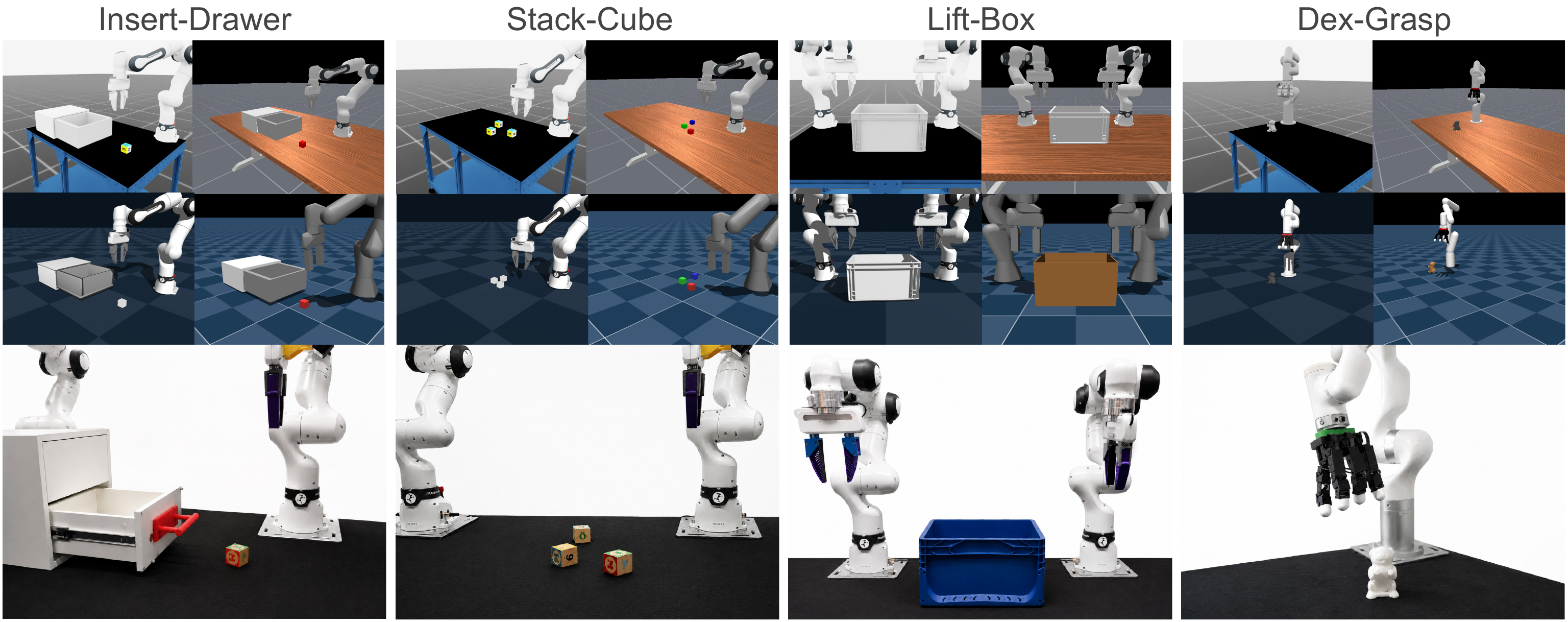}
    \vspace{-1mm}
    \caption{
    End-to-end policy learning across four tasks and four simulators: IsaacLab, ManiSkill, Genesis, and MJLab.
    Simulator snapshots on top and the corresponding real-world setup below.
    }
    \vspace{-5mm}
    \label{fig:tasks}
\end{figure}

% Built on this suite, our experiments 
We evaluate HARBOR along three axes: (i) we demonstrate that HARBOR can support an end-to-end simulation workflow with real-world validation in Sec.~\ref{sec:end-to-end}; (ii) we evaluate its ability to integrate and tune RL algorithms across diverse benchmarks in Sec.~\ref{sec:tuning}; (iii) we provide ablations and efficiency analyses that quantify the effect of key design choices, wall-clock time, and token cost. We comprehensively evaluate HARBOR across 6 benchmarks, each with a different simulation backend, including IsaacLab~\citep{mittal2025isaac}, Bi-DexHands(IsaacGym)~\citep{chen2022towards, makoviychuk2021isaac}, ManiSkill(Sapien)~\citep{gu2023maniskill2, xiang2020sapien}, Genesis~\citep{genesis2026genesisworld}, MJLab(Warp)~\citep{zakka2026mjlablightweightframeworkgpuaccelerated}, and Loco-Mujoco(MJX)~\citep{alhafez2023b, todorov2012mujoco}. This diverse evaluation suite is designed to stress-test the generality and flexibility of HARBOR across heterogeneous robot-learning settings. For all simulation experiments, we train with five random seeds and evaluate each seed using $4{,}096$ rollouts, reporting success rates or cumulative returns as mean and standard error. Additional learning curves are provided in Appx.~\ref{app:curves}. For real-world experiments, each policy is evaluated over 10 trials; the hardware setup is described in Appx.~\ref{app:real_setup}.

% \paragraph{Simulators and tasks}
% We conduct experiments on six mainstream simulators: IsaacLab, IsaacGym, ManiSkill/SAPIEN, Genesis-World, MJLab, and MJX. These platforms cover many of the most widely used simulation ecosystems in robot learning and expose heterogeneous training backends, including PyTorch- and JAX-based workflows. This setup allows us to evaluate whether HARBOR can adapt not only across simulators and task APIs, but also across different computational backends. For the end-to-end sim-to-real experiments, we evaluate HARBOR on four manipulation tasks as shown in Fig.~\ref{fig:tasks}: \texttt{stack-cube}, \texttt{insert-drawer}, \texttt{lift-box}, and \texttt{dex-grasping}. These tasks cover long-horizon composition, interaction with articulated objects, bimanual coordination, and dexterous control. For RL tuning, we select four table-top manipulation tasks from IsaacLab, four bimanual dexterous manipulation tasks from Bi-DexHands, and four locomotion tasks from Loco-MuJoCo. Together, these experiments provide broad coverage over representative simulators, task families, and training pipelines, enabling a systematic evaluation of HARBOR's usability and scalability. 

\subsection{Automated End-to-End Simulation Pipeline and Real-world Validation}
\label{sec:end-to-end}
\paragraph{Automated End-to-End Simulation Pipeline}
We first evaluate whether HARBOR can automate robot-RL workflows in simulation across different simulators and tasks. We select four testbeds, IsaacLab, ManiSkill, Genesis, and MJLab, and implement four manipulation tasks in each: \texttt{stack-cube}, \texttt{insert-drawer}, \texttt{lift-box}, and \texttt{dex-grasping}. These tasks cover long-horizon composition, articulated-object interaction, bimanual coordination, and dexterous control. Given a task specification containing scene assets, success metrics, and desired behaviors (Appx.~\ref{app:task_prompt}), HARBOR performs dependency setup, task implementation, reward generation, RL integration, and policy training. As shown in Fig.~\ref{fig:tasks}, it produces semantically consistent task instances despite simulator-specific APIs, asset formats, and contact interfaces. 

As shown in Tab.~\ref{tab:sim2real_reward_eval}, HARBOR achieves strong performance on IsaacLab, ManiSkill, and Genesis, reaching $88.5\% \sim 100\%$ success on all four IsaacLab tasks, including three-cube stacking, a long-horizon variant beyond the standard two-cube benchmark. Performance is lower on MJLab, especially for dexterous-hand control. Based on the automation logs, we attribute these failures primarily to simulator physics: HARBOR struggles to identify physics parameters that yield stable robot control in MJLab. In contrast, on the other simulators, HARBOR successfully adapts to different simulation dynamics and produces similar high-level reward structures, suggesting that it can preserve task and reward intent while adjusting simulator-specific implementation details.

\begin{table*}[h]
\centering
\small
\setlength{\tabcolsep}{8.0pt}
\renewcommand{\arraystretch}{1.0}
\resizebox{\textwidth}{!}{
\begin{tabular}{c|ccc|c|cc|cc|cc}
\toprule
\multirow{3}{*}{\textbf{Task}}
& \multicolumn{4}{c|}{\textbf{IsaacLab}}
& \multicolumn{2}{c|}{\textbf{ManiSkill}}
& \multicolumn{2}{c|}{\textbf{Genesis}}
& \multicolumn{2}{c}{\textbf{MJLab}} \\
\cmidrule(lr){2-5}
\cmidrule(lr){6-7}
\cmidrule(lr){8-9}
\cmidrule(lr){10-11}

& \multicolumn{3}{c|}{\textbf{Sim}}
& \multirow{2}{*}{\textbf{Real}}
& \multirow{2}{*}{\textbf{Sim}}
& \multirow{2}{*}{\textbf{Real}}
& \multirow{2}{*}{\textbf{Sim}}
& \multirow{2}{*}{\textbf{Real}}
& \multirow{2}{*}{\textbf{Sim}}
& \multirow{2}{*}{\textbf{Real}} \\
\cmidrule(lr){2-4}

& \textbf{Ours}
& \textbf{Eureka}
& \textbf{REvolve}
& Ours
& Ours
& Ours
& Ours
& Ours
& Ours
& Ours\\
\midrule

\texttt{Stack-Cube}
& 0.885 & 0.0 & 0.0 & 0.5
& \textbf{0.936} & \textbf{0.6}
& 0.935 & 0.6
& 0.221 & 0.1 \\

\texttt{Insert-Drawer}
& \textbf{0.939} & 0.0 & 0.363 & \textbf{0.6}
& 0.648 & 0.4
& 0.781 & 0.4
& 0.122 & 0.0 \\

\texttt{Lift-Box}
& \textbf{1.0} & 0.997 & 0.932 & \textbf{0.8}
& \textbf{1.0} & \textbf{0.8}
& 0.944 & \textbf{0.8}
& \textbf{1.0} & 0.2 \\

\texttt{Dex-Grasp}
& 0.967 & 0.935 & 0.955 & \textbf{0.9}
& 0.932 & 0.7
& \textbf{0.970} & 0.7
& 0.0 & 0.0 \\

\bottomrule
\end{tabular}
}
\caption{
End-to-end simulation results and real-world validations across four simulators and tasks.
}
\label{tab:sim2real_reward_eval}
\vspace{-3mm}
\end{table*}

\paragraph{Reward Design Comparison} We isolate automated reward design by comparing HARBOR with two LLM-based reward-generation baselines, Eureka~\citep{ma2024eureka} and REvolve~\citep{hazra2025revolve}, on IsaacLab. For a fair comparison, all methods use the same Opus backbone, wall-clock budget, task specification, and fixed task implementation, including observations, actions, and reset logic. As shown in Tab.~\ref{tab:sim2real_reward_eval}, HARBOR consistently outperforms both baselines across tasks due to two major factors. (1) Rollout-behavior feedback is essential: in \texttt{insert-drawer}, both HARBOR and REvolve identify failures such as insufficient object lifting and collisions with the drawer, whereas Eureka fails to diagnose them and gets stuck; (2) pre-provided human heuristics are important: for the long-horizon \texttt{stack-cube} task, HARBOR succeeds by using relevant heuristics to construct staged rewards, where each stage is activated only after the previous one is completed; in contrast, Eureka and REvolve search from scratch and fail. Finally, we test whether accumulated experience improves reward design by having HARBOR redesign the reward in a separate run. Using prior experience, it reduces wall-clock time from 4 hours to 30 minutes, an $8\times$ speedup on the hardest \texttt{stack-cube}.
\vspace{-2mm}
\paragraph{Real-world Validation} We validate sim-to-real transfer on real robots for all four tasks, using policies trained in different simulators. HARBOR supports transfer through system identification and domain randomization: given user-provided real-world trajectories, it searches for simulator physics parameters that best match real rollouts, and adjusts randomization ranges such as object mass and initial object pose using human feedback. As shown in Tab.~\ref{tab:sim2real_reward_eval}, policies produced by this pipeline successfully transfer to real-world execution across the evaluated tasks, with moderate success rates under simulator-specific dynamics and contact differences.

\subsection{RL Algorithm Tuning}
\label{sec:tuning}
\begin{figure}[t!]
    \centering
    \includegraphics[width=\textwidth]{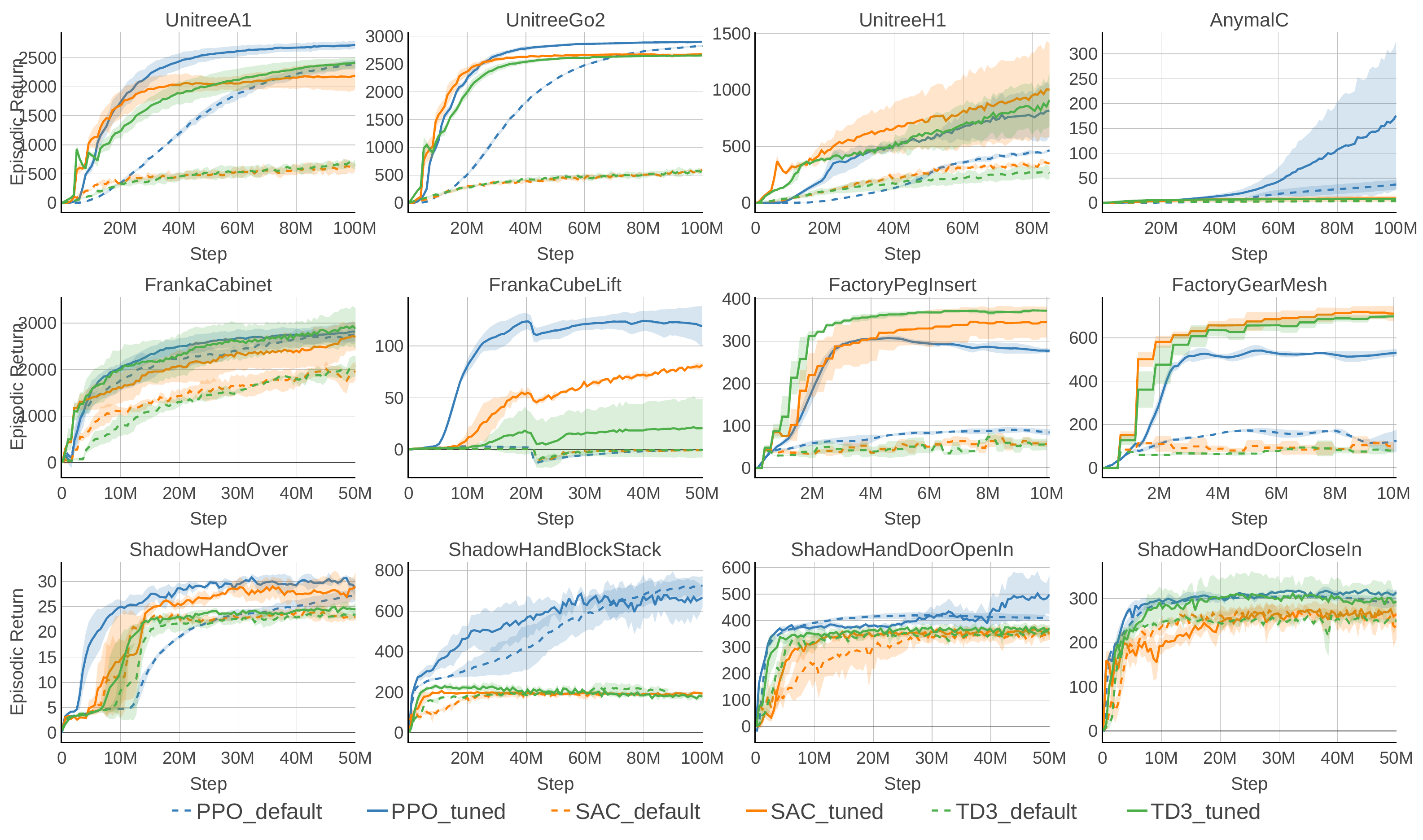}
    \vspace{-6mm}
    \caption{Performance of HARBOR on RL tuning across three benchmarks, each with four tasks. HARBOR generally improves the performance (solid) over defaults (dashed).}
    \label{fig:rl-tuning}
    \vspace{-5mm}
\end{figure}

We test whether HARBOR can tune standard RL algorithms, including PPO~\citep{schulman2017proximal}, SAC~\citep{haarnoja2018soft}, and TD3~\citep{fujimoto2018addressing}, under a bounded wall-clock budget. We use four table-top manipulation tasks from IsaacLab, four bimanual dexterous manipulation tasks from Bi-DexHands, and four locomotion tasks from Loco-MuJoCo. The default PPO configurations are taken from IsaacLab, while SAC and TD3 defaults are taken from PQL~\citep{li2023parallel} (dashed lines in Fig.~\ref{fig:rl-tuning}). HARBOR launches parallel tuning trials and selects configurations whose training time is at most twice that of the corresponding default. HARBOR applies common stabilization techniques, such as observation and reward normalization, and adjusts algorithm-specific hyperparameters using retrieved heuristics and accumulated tuning experience. As shown in Fig.~\ref{fig:rl-tuning}, the tuned configurations match or outperform the defaults in 11 of 12 task settings; detailed sample-efficiency metrics are reported in Appx.~\ref{app:rl_tuning_table}. The gains are most notable for SAC on Bi-DexHands, where HARBOR-tuned SAC learns \texttt{ShadowHandDoorOpenIn} and \texttt{ShadowHandDoorCloseIn}, tasks on which the original SAC baseline fails as in its paper.

\subsection{Ablations and Efficiency Analysis}
\label{sec:ablation}

We ablate HARBOR on a clean ManiSkill checkout by building and training a policy for the \texttt{Push-Cube} task as a controlled profiling slice. 
We compare five configurations: the full harness, three ablations (\emph{w/o CCDE}, removing subagent isolation and parallel execution; \emph{w/o gate}, removing validation gates; \emph{w/o experience}, removing templates, references, and heuristics), and a no-harness \emph{Vanilla} baseline. 
Each configuration is repeated ten times. 
For each stage, we report success rate (S.R., out of 10), wall-clock time, context tax (Ctx, cache-read tokens re-read across turns), and estimated cost in Tab.~\ref{tab:workflow_ablation_transposed_clean}. 
Detailed stage definitions and profiling protocols are provided in Appx.~\ref{app:ablation}.

The ablation reveals clear reliability and cost differences, especially in iteration-heavy stages such as reward generation and RL tuning. 
First, isolation improves both efficiency and reliability. 
By delegating trials to isolated subagents and running reward/tuning attempts in parallel, the full harness reduces the iteration-heavy stages to $7.9$\,min, compared with $49.5$\,min for \emph{w/o CCDE} (${\approx}6.3\times$ faster). Without subagent isolation, transcript re-reads grow substantially ($65.3$M vs.\ $28.07$M Ctx), serial trials dominate wall-clock time, and reward/tuning success drops to $8/10$. 
Second, gates catch silent failures. The \emph{w/o gate} variant is fast but allows a render-path defect to pass unnoticed (marked $^\dagger$) and achieves only $6/10$ success on both reward generation and RL tuning. 
Third, experience improves generative stages. Removing templates, references, and heuristics makes \emph{w/o experience} the least reliable on reward generation ($2/10$), RL tuning ($4/10$), and RL integration ($8/10$). Overall, HARBOR achieves the highest reliability ($48/50$) while also being the cheapest measured harness configuration, making it Pareto-optimal on reliability and cost.

% Beyond compute, the main benefit is reduced human attention. 
% A fresh sim-to-real bring-up can require $8$--$25$ hours of expert intervention across setup, debugging, reward iteration, and tuning. 
% HARBOR reduces this to roughly $12$ minutes of interaction in the reference run, while using comparable GPU trials. 
% The speedup therefore comes from removing the manual labor around training, not from making training itself cheaper.

\begin{table*}[t]
\centering
\footnotesize
\setlength{\tabcolsep}{3.6pt}
\renewcommand{\arraystretch}{1.18}
\caption{
Ablations on the ManiSkill \texttt{Push-Cube} pipeline. $^{\dagger}$ marks a stage that reported success but hid a silent render-path defect. We use default $N=4$ for parallelization and Opus 4.7, while results for Sonnet is in Appx.~\ref{app:sonnet}.
}
\label{tab:workflow_ablation_transposed_clean}
\resizebox{\textwidth}{!}{
\begin{tabular}{
>{\centering\arraybackslash}p{2.7cm}
ccc
ccc
ccc
ccc
ccc
}
\toprule
\textbf{Stage}
& \multicolumn{3}{c}{Vanilla}
& \multicolumn{3}{c}{\textsc{Harbor}}
& \multicolumn{3}{c}{w/o CCDE}
& \multicolumn{3}{c}{w/o gate}
& \multicolumn{3}{c}{w/o experience} \\
\cmidrule(lr){2-4}
\cmidrule(lr){5-7}
\cmidrule(lr){8-10}
\cmidrule(lr){11-13}
\cmidrule(lr){14-16}
& S.R. & Time & Ctx/\$
& S.R. & Time & Ctx/\$
& S.R. & Time & Ctx/\$
& S.R. & Time & Ctx/\$
& S.R. & Time & Ctx/\$ \\
\midrule
Dependency setup
& 10/10 & 21.86 & 4.78M / \$17.91
& 10/10 & 11.2 & 2.78M / \$8.20
& 10/10 & 12.4 & 4.36M / \$15.35
& 10/10 & 13.1 & 2.10M / \$12.74
& 10/10 & 10.6 & 0.68M / \$3.52 \\

RL integration
& 3/10 & 29.73 & 15.16M / \$31.42
& 10/10 & 17.1 & 3.72M / \$9.12
& 10/10 & 17.6 & 12.1M / \$24.64
& 10/10$^{\dagger}$ & 4.1 & 1.29M / \$3.52
& 8/10 & 14.8 & 7.06M / \$13.81 \\

Task generation
& 8/10 & 13.78 & 5.28M / \$12.77
& 10/10 & 7.9 & 2.92M / \$8.38
& 10/10 & 3.9 & 3.99M / \$8.09
& 9/10 & 5.2 & 1.13M / \$4.04
& 8/10 & 10.5 & 1.90M / \$6.54 \\

Reward generation
& 3/10 & 49.20 & 25.65M / \$48.10
& 9/10 & 4.2 & 14.35M / \$31.00
& 8/10 & 24.0 & 24.80M / \$44.45
& 6/10 & 2.9 & 7.05M / \$17.90
& 2/10 & 6.0 & 10.25M / \$29.05 \\

RL tuning
& 5/10 & 71.05 & 31.90M / \$71.95
& 9/10 & 3.7 & 4.30M / \$14.20
& 8/10 & 25.5 & 20.05M / \$38.15
& 6/10 & 7.6 & 12.40M / \$41.50
& 4/10 & 11.6 & 24.50M / \$67.35 \\

\midrule
\textbf{Total ($\Sigma$-win.)}
& 29/50 & 185.62 & 82.77M / \$182.15
& 48/50 & 44.1 & 28.07M / \$70.90
& 46/50 & 83.4 & 65.30M / \$130.68
& 41/50$^{\dagger}$ & 32.9 & 23.97M / \$79.70
& 32/50 & 53.5 & 44.39M / \$120.27 \\
\bottomrule
\end{tabular}
}
\vspace{-5mm}
\end{table*}
%===============================================================================
% \vspace{-3mm}
\section{Conclusion, Limitations, and Future Work}
\label{sec:conclusion}
We presented \textbf{HARBOR}, a harness framework for automating robot RL workflows with LLM agents. HARBOR is motivated by the structure of robot RL automation, where MDPs provide stable interfaces and rollout behaviors provide executable feedback for validation. It turns natural-language requests into auditable workflows through agents, commands, artifacts, gates, and knowledge. Beyond this abstraction, HARBOR supports asynchronous parallel execution and experience reuse, enabling the system to scale and improve iterative stages. We implement HARBOR as an LLM-agent plugin and evaluate it across diverse simulators and robot-learning tasks, which shows that HARBOR can automate end-to-end RL workflows in simulation, tune RL algorithms, and reduce execution cost and wall-clock time compared with vanilla LLM agents.

\paragraph{Limitations and Future Work}
HARBOR's capability is \textbf{bounded by the tools exposed to its agents}: for novel tasks without prior knowledge, it may require many scaffold-and-repair iterations or fail outright. As more users apply HARBOR to new scenarios, the shared knowledge base can be expanded by distilling from successful workflows, heuristics, and failure diagnoses. A second limitation is the \textbf{sim-to-real boundary}: HARBOR automates much of the simulated pipeline, but deployment to physical robots still requires manual engineering around robot interfaces and human feedback. Future versions could extend the harness to the real-robot side, closing the loop for sim-real alignment. Finally, HARBOR currently supports relatively simple policy architectures; \textbf{extending to vision-language-action or world-model-based policies} is an important direction, and HARBOR's abstraction makes such extensions additive rather than requiring a system redesign.

\clearpage
% The acknowledgments are automatically included only in the final and preprint versions of the paper.
\acknowledgments{This research work has received funding from the Emmy Noether Programme DFG Project Nr. CH 2676/1-1, Federal Ministry of Research, Technology and Space (BMFTR) Project “RIG” (Grant No.: 16ME1001), and the Project Pose Confidences for Telerobotics from Honda Research Institute, Europe.}

%===============================================================================

\bibliography{main}  % .bib

%===============================================================================
\newpage
%===============================================================================
\appendix
\section{Related Work}
\label{app:related}

\paragraph{LLM-driven reward engineering and sim-to-real transfer}
Recent work has shown that large language models (LLMs) can substantially reduce human effort in reward engineering and sim-to-real transfer. Eureka~\citep{ma2024eureka} uses LLMs to generate and iteratively improve executable reward functions; Text2Reward~\citep{xie2024text2reward} similarly converts natural-language task descriptions into dense reward code for manipulation and locomotion; and REvolve~\citep{hazra2025revolve} further studies reward evolution with human feedback, using LLMs as mutation and refinement operators to incorporate implicit human preferences into reward design. Beyond reward engineering, Dr.~Eureka~\citep{ma2024dreureka} extends this paradigm to sim-to-real transfer by jointly producing reward functions and domain-randomization distributions, while DexSim2Real~\citep{zeng2026dexsim2real} explores LLM-assisted design for dexterous sim-to-real policy learning. These methods establish LLMs as effective tools for automating individual components of robot RL. In contrast, HARBOR targets the broader end-to-end RL workflow: given a simulator codebase and task specification, it automates from environment setup, task integration, reward and domain-randomization design, to policy training.

\paragraph{AutoRL and hyper-parameter tuning}
HARBOR is closely related to automated reinforcement learning (AutoRL), especially the automation of algorithm and hyper-parameter tuning~\citep{parker2022automated, afshar2022automated}. Classical hyperparameter-optimization methods such as random search~\citep{bergstra2012random}, Bayesian optimization~\citep{wu2019hyperparameter}, Hyperband~\citep{li2018hyperband}, and population-based training~\citep{jaderberg2017population, liu2021elegantrl} provide general mechanisms for efficient configuration search. Practical systems such as Ray Tune~\citep{liaw2018tune} and Optuna~\citep{akiba2019optuna} make these methods accessible at scale and have become standard tools for distributed tuning. However, most AutoRL and HPO frameworks assume that the training environment, logging, evaluation scripts, and search space are already specified by humans. HARBOR is complementary to these systems: it not only supports adaptive and distributed tuning, but also automates the surrounding engineering process, enabling robot RL experiments to reach a higher level of automation.

\paragraph{Agentic systems and harnesses}
Recent LLM agents have shown strong potential for automating complex workflows by decomposing tasks, calling tools, maintaining state, and refining outputs across multiple steps~\citep{yang2024swe, jiang2026survey, steinberger_openclaw}. A growing view is that their capability comes not only from the underlying LLM, but also from the surrounding harness, including tools, memory, artifacts, execution environments, and feedback channels~\citep{meng2026agent,he2026harness,zhou2026externalization}. Meta-harnesses further aim to standardize and automate these interfaces for long-horizon agentic workflows~\citep{he2026harness}. The closest concurrent work to HARBOR is Nautilus~\citep{jin2026nautilus}, which automates imitation learning policy evaluation and benchmarking, especially for vision-language-action (VLA) models. In contrast, HARBOR targets robot RL workflows, where automation requires simulator integration, MDP specification, reward and domain-randomization design,  and policy training.

\paragraph{Sim-to-real reinforcement learning for robotics}
Sim-to-real reinforcement learning has emerged as a promising paradigm for acquiring complex robotic behaviors, with successful demonstrations ranging from agile locomotion~\citep{margolis2024rapid,li2025reinforcement} to dexterous manipulation such as object reorientation~\citep{chen2023visual}, lid twisting~\citep{lin2024twisting}, and even more challenging bimanual dexterous manipulation tasks~\citep{li2025morphologically}. Recent results further show that sim-to-real RL can handle highly dynamic tasks such as table tennis~\citep{su2025hitter} and badminton~\citep{ma2025learning}, suggesting its potential as a scalable recipe for robot skill acquisition. However, these successes typically rely on substantial task-specific engineering, including simulator construction, task and MDP design, reward shaping, domain randomization. HARBOR is motivated by this bottleneck: it aims to automate the engineering loop needed to rapidly train and deploy RL policies for individual robot tasks, while laying the groundwork for scaling sim-to-real RL across many tasks toward more generalist robot policies.

%===============================================================================
\section{Background: Harness Engineering}
\label{app:background}

A harness is the external structure that enables an LLM agent to perform reliable work in a complex environment. It is more than a prompt: it specifies the instructions, tools, runtime environment, persistent state, and feedback channels through which the LLM agent operates. Recent discussions emphasize that a harness does not make the model itself smarter, but instead turns the model into part of a closed-loop working system~\citep{lopopolo2026harness, rajasekaran2026harness}. In this view, agent performance depends not only on the underlying LLM, but also on whether the surrounding system makes the task legible, executable, and verifiable.

\paragraph{Why harnesses are needed}
Long-horizon agentic tasks require the model to make many dependent decisions over time, where early mistakes or missing context can compound into later failures. Human engineers manage such workflows with scaffolding such as setup scripts, documentation, tests, logs, progress records, and clear interfaces; agents need analogous structure to operate reliably. OpenAI describes this shift as moving from manually writing code to designing environments, specifying intent, and building feedback loops that allow agents to execute reliable work~\citep{lopopolo2026harness}. Without such structure, agents may lose context, repeat failed attempts, overfit to local instructions, or declare success before the full task has been validated.

\paragraph{Core components}
A practical harness consists of several interacting components~\citep{meng2026agent}. 
\begin{itemize}[leftmargin=*]
    \item The \textit{instruction} layer gives the agent task goals, project conventions, constraints, and pointers to relevant documents. 
    \item The \textit{tool} layer exposes executable actions, such as file editing, shell commands, tests, simulators, logging systems, or deployment scripts. 
    \item The \textit{environment} layer defines dependencies, containers, and runtime versions. 
    \item The \textit{state} layer records plans, completed steps, failures, and intermediate artifacts. 
    \item The \textit{feedback} layer provides executable checks, such as unit tests, integration tests, evaluation metrics, rollout diagnostics, or human review.
\end{itemize}
These components are instantiated in HARBOR and specialized for robot RL. For example, the feedback layer is implemented as executable gates that verify each workflow stage before proceeding. HARBOR also extends the standard harness design with robot-RL-specific mechanisms such as reusable experiences, which store prior reward-design, domain-randomization, and tuning traces as continual guidance for future tasks.

\paragraph{Harness implementation}
A key principle in implementing a harness is that information invisible to the agent is effectively unavailable. Therefore, a practical harness usually needs both top-level instructions and low-level details. The top-level instructions define the overall goal, workflow structure, constraints, and where the agent should look next, while the low-level details provide the concrete references, templates, scripts, and artifacts needed to execute each step. This separation avoids placing all information in one giant prompt, but still ensures that the agent can access the knowledge required for reliable execution~\citep{rajasekaran2026harness}. HARBOR follows this design: its agents and commands explicitly specify which references, templates, scripts, and artifacts should be used at each workflow stage. In this way, the top-level instruction acts as a map, while detailed knowledge is distilled from existing codes and organized in structured files.

%===============================================================================
\section{List of Agents and Commands}
\label{app:agent_list}

%========================= 1. Dependency Agent =========================
\subsection{Dependency-generator (\textit{Stage: Dependency generation/package installation})}
\begin{itemize}[leftmargin=*]
    \item \textbf{Function:}
    \begin{enumerate}[leftmargin=*, nosep]
        \item Probe the repository's dependency surface, read the \texttt{README.md} file, and render an installation plan \texttt{install\_plan.json}. 
        \item Render a reproducible \texttt{setup\_uv.sh} that creates an isolated \texttt{.venv/} via uv.
        \item Render a \texttt{setup\_uv.sh} that contains instructions on how to activate the virtual environment.
    \end{enumerate}
    \item \textbf{Commands:} \texttt{/harbor:env-install-uv}.
    \item \textbf{Gated test:}
    \begin{enumerate}[leftmargin=*, nosep]
        \item Basic environment such as CUDA is safe and detectable.
        \item The \texttt{.venv/} is created and the editable package imports cleanly.
    \end{enumerate}
\end{itemize}

%========================= 2. Benchmark Sanity Agent =========================
\subsection{\textbf{Benchmark-generator (\textit{Stage: Benchmark scaffolding})}}
\begin{itemize}[leftmargin=*]
    \item \textbf{Function:}
    \begin{enumerate}[leftmargin=*, nosep]
        \item Probe the repository and summarize the benchmark in \texttt{benchmark-spec.json}, such as the backend (Jax or Torch), support for GPU-based simulation. 
        \item Author a \texttt{task\_overview.md} that summarizes the task distribution in the benchmark and details of each task, including description, reward implementation, obs. and action space. 
        \item Author a \texttt{task\_implementation.md} that provides details on APIs for task creation using \texttt{/harbor:probe-benchmark}.
        \item Render the random-action rollout and render-to-MP4 entry scripts against the benchmark.
    \end{enumerate}
    \item \textbf{Commands:} \texttt{/harbor:probe-benchmark}.
    \item \textbf{Gated test:}
    \begin{enumerate}[leftmargin=*, nosep]
        \item Prerequisite check on if the dependency-generator's uv environment is healthy.
        \item The random-action rollout runs without error and gets finite rewards.
        \item The render pass produces a valid MP4 video.
        \item The suite spec is captured with a non-empty task list.
    \end{enumerate}
\end{itemize}

%========================= 3. Task Authoring Agent =========================
\subsection{\textbf{Task-generator (\textit{Stage: Task generation})}}
\begin{itemize}[leftmargin=*]
    \item \textbf{Function:}
    \begin{enumerate}[leftmargin=*, nosep]
        \item Search if there exists tasks in experiences that are relevant or similar to the desired task.
        \item Author task registration and scene, action terms, reset, goal and termination, and observation.
        \item Leave a placeholder reward and an empty domain-randomization slot so the environment still builds for downstream stages.
        \item Render a \texttt{task-history.md} to document all design choices and test results.
    \end{enumerate}
    \item \textbf{Commands:} \texttt{/harbor:task-create}, \texttt{/harbor:probe-task}, \texttt{/harbor:rl-render}.
    \item \textbf{Gated test:}
    \begin{enumerate}[leftmargin=*, nosep]
        \item The task is registered in the benchmark and can be initialized.
        \item The action target matches the expected value in controller and can effectively control the robot.
        \item The reset values match the simulation state; and the goal appears in the observation and a forced termination fires.
        \item The observation order, shapes, and values pass.
    \end{enumerate}
\end{itemize}

%========================= 4. Reward Authoring Agent =========================
\subsection{\textbf{Reward-generator (\textit{Stage: Reward generation})}}

\begin{itemize}[leftmargin=*]
    \item \textbf{Function:}
    \begin{enumerate}[leftmargin=*, nosep]
        \item Support both creating and editing mode that can either write the rewards from scratch or modify the existing rewards.
        \item Search the experience for a similar task's reward and author the reward (term ladder, weights, composer, stage gating) with reward term log.
        \item Author \texttt{reward-history.md} for iteration history and \texttt{handoff-reward-generator.md} for latest reward terms tracking.
        \item Render and run the reward smoke; optionally iterate with training in the loop until success.
    \end{enumerate}
    \item \textbf{Commands:} \texttt{/harbor:reward-tune}, \texttt{/harbor:reward-add-log}, \texttt{/harbor:rl-run}, \texttt{/harbor:rl-render}.
    \item \textbf{Gated test:}
    \begin{enumerate}[leftmargin=*, nosep]
        \item The reward is finite and matches expected across a rollout.
        \item The composer over the per-term values equals the environment reward at every step (sum or product).
        \item Under tuning, the trained policy's success rate reaches the target threshold.
    \end{enumerate}
\end{itemize}

%========================= 5. Domain Randomization Agent =========================
\subsection{\textbf{DR-generator (\textit{Stage: Domain randomization generation})}}
\begin{itemize}[leftmargin=*]
    \item \textbf{Function:}
    \begin{enumerate}[leftmargin=*, nosep]
        \item Discover every available randomization term across the robot, object, and observation groups.
        \item Wire each term once per episode per environment (multiplicative / additive / direct for physical properties; uniform / gaussian for observation noise).
        \item Render and run the exact-value randomization smoke.
    \end{enumerate}
    \item \textbf{Commands:} \texttt{/harbor:task-create}, \texttt{/harbor:rl-run}, \texttt{/harbor:rl-sweep}.
    \item \textbf{Gated test:}
    \begin{enumerate}[leftmargin=*, nosep]
        \item Each effective term's value equals the default value modified by a known sampled value.
        \item The randomization re-applies correctly after a reset.
        \item Observation-noise terms diverge from a paired no-noise rollout at the configured magnitude.
    \end{enumerate}
\end{itemize}

%========================= 6. RL Integration Agent =========================
\subsection{\textbf{RL-integration-generator (\textit{Stage: RL algorithm integration})}}
\begin{itemize}[leftmargin=*]
    \item \textbf{Function:}
    \begin{enumerate}[leftmargin=*, nosep]
        \item Implement default RL algorithms (PPO, SAC, TD3) or user-provided RL implementation with per-algorithm training, evaluation, and render scripts, the configs, and the DataLogger.
        \item Wire the implementation and canonical metric-logging contract with the benchmark.
    \end{enumerate}
    \item \textbf{Commands:} \texttt{/harbor:rl-run}, \texttt{/harbor:rl-render}, \texttt{/harbor:rl-add-log}.
    \item \textbf{Gated test:}
    \begin{enumerate}[leftmargin=*, nosep]
        \item A training step runs and evaluation writes \texttt{metrics.json}.
        \item The render produces a moving-policy MP4 (frame-difference check).
        \item The learning curves are plotted and the metric keys are present in TensorBoard / W\&B.
    \end{enumerate}
\end{itemize}

%========================= 7. RL Tuning Agent =========================
\subsection{\textbf{RL-tuning-agent (\textit{Stage: Hyper-parameter tuning})}}
\begin{itemize}[leftmargin=*]
    \item \textbf{Function:}
    \begin{enumerate}[leftmargin=*, nosep]
        \item Run an open-ended per-(task, algorithm) tuning loop: default-config baseline, then training tricks, then log-driven hyperparameter edits.
        \item Train, evaluate, render, and analyze metrics and behavior each iteration to propose the next configuration.
        \item Stop when the running best is not beaten for a fixed number of consecutive iterations and emit a convergence plot.
    \end{enumerate}
    \item \textbf{Commands:} \texttt{/harbor:rl-tune}, \texttt{/harbor:rl-sweep}, \texttt{/harbor:rl-add-trick}, \texttt{/harbor:rl-list-tricks}.
    \item \textbf{Gated test:}
    \begin{enumerate}[leftmargin=*, nosep]
        \item Each trial trains and writes the canonical metrics.
        \item Evaluation yields an unbiased steady-state aggregate and the render confirms motion.
        \item The tuned configuration improves sample efficiency and final return over the default baseline.
    \end{enumerate}
\end{itemize}
%===============================================================================

\section{Experiments Details}

\subsection{Real-world Experiment Setup}
\label{app:real_setup}

\paragraph{Robot Platforms, Sensors, and Control}
We evaluate real-world deployment on two robot platforms. The first platform is a bimanual Franka setup consisting of two 7-degree of freedom(DoF) Franka arms, each equipped with a custom 3D-printed parallel gripper, yielding a 14-DoF arm system plus two gripper actuation channels. The second platform is a single 6-DoF xArm UF850 arm equipped with a 16-DoF Allegro Hand V4. 
% Both setups are built on the Nautilus data-collection framework~\citep{jin2026nautilus} and operate on the same tabletop workspace. 
Each setup uses a single external ZED2 RGB-D camera for visual perception; we do not use any wrist-mounted or additional cameras. We use FoundationPose~\citep{wen2024foundationpose} and SAM2.1~\citep{ravi2024sam} to obtain object masks and 6D object poses for closed-loop policy execution.

For the bimanual Franka setup, policies run at 20Hz and output end-effector delta-pose actions for each arm. For the xArm-Allegro setup, policies also run at 20Hz, but output delta joint-position actions. The xArm low-level joint controller runs at 120Hz, while the Allegro Hand receives corresponding joint-position targets. For both platforms, we apply exponential moving average (ema) smoothing to the policy actions before sending them to the robot controllers, which reduces high-frequency action noise during sim-to-real execution. Specifically, the command action for every step is computed as follows: $a^\text{cmd}_{t+1} = a^\text{cmd}_t + \eta \cdot \text{EMA}(a_t)$, where $\eta$ is a scaling factor.

% \section{Perception}\label{app:perception}
\paragraph{Perception Pipeline} Our perception pipeline combines FoundationPose~\cite{wen2024foundationpose} and SAM2.1~\cite{ravi2024sam} to achieve robust, real-time 6D object pose tracking in cluttered and dynamic scenes. The input consists of RGB-D frames captured at 1080p and 30 FPS from a single external ZED2 camera. We operate the camera in ultra mode to maximize depth range and preserve Z-accuracy along the sensing axis, which is crucial for high-precision pose estimation.

We use FoundationPose in a tracking-first mode rather than running full registration on every frame. At the beginning of a track, or after tracking failure, the registration module provides an initial 6D pose hypothesis. During normal execution, we avoid repeated registration and instead initialize the refiner with a coarse pose prior estimated from the SAM2.1 mask centroid and mean masked depth. The refiner then updates the pose from the current RGB-D frame. This design keeps the perception loop efficient enough for closed-loop policy execution while preserving the ability to reinitialize when tracking is lost. We use SAM2.1 for multi-object instance segmentation and tracking, and for each incoming frame the refiner is executed in parallel for known objects to predict their 6D poses. 

While FoundationPose is robust under typical conditions and performs well on standard benchmarks, it fails to recover object pose when faced with rapid motion or complete occlusion. To handle such cases, we run a lightweight pose-quality evaluation at 2 FPS to detect tracking failures without slowing down the main tracking loop. For each object we render expected RGB and depth images using a lightweight offscreen renderer based on the object’s CAD model. These rendered views are compared against the observed images from the ZED2 camera, and the rendered-vs-observed depth agreement gives a pose likelihood. Pose confidence is computed by measuring photometric and geometric discrepancies between the rendered and observed RGB-D images. Specifically, we define the confidence score as:
% \begin{equation}
% c = \exp\left( -\frac{1}{N} \sum_{i=1}^N \left( \left\| I_{\text{obs}}^{(i)} - I_{\text{rend}}^{(i)} \right\|_1 + \lambda \cdot \left\| D_{\text{obs}}^{(i)} - D_{\text{rend}}^{(i)} \right\|_1 \right) \right)
% \end{equation}
\begin{equation}
	c = \exp\left( -\frac{1}{\sum_{i} M^{(i)}} \sum_{i} M^{(i)} \left( \left\| I_{\text{obs}}^{(i)} - I_{\text{rend}}^{(i)} \right\|_1 + \lambda \cdot \left\| D_{\text{obs}}^{(i)} - D_{\text{rend}}^{(i)} \right\|_1 \right) \right)
\end{equation}

where $M^{(i)}$ is a binary foreground mask obtained from SAM2.1. $I_{\text{obs}}$, $I_{\text{rend}}$ denote observed and rendered RGB images, $D_{\text{obs}}$, $D_{\text{rend}}$ denote depth images, and $\lambda$ balances color and depth contributions. If the confidence $c < 0.5$, the object is deemed lost, and its pose is re-initialized by re-running FoundationPose's registration network, without relying on temporal priors.

To mitigate jitter and ensure smooth input to the policy, we apply SLERP interpolation~\cite{kremer2008quaternions} for rotations and linear interpolation for translations in SE(3) across consecutive pose estimates, followed by exponential moving average (EMA) filtering. This ensures temporally coherent trajectories and stabilizes the pose stream at a steady 15 FPS for downstream policy execution.

\subsection{Task Prompts for End-to-End Task Generation}
\label{app:task_prompt}

\begin{tcolorbox}[
    colback=blue!2,
    colframe=blue!20,
    boxrule=0.5pt,
    arc=2pt,
    left=6pt,
    right=6pt,
    top=5pt,
    bottom=5pt,
    fontupper=\ttfamily\small,
    title={\textnormal{\textsc{Insert-drawer} (IsaacLab example)}},
    coltitle=black,
    colbacktitle=blue!6
]
/harbor:task-create In IsaacLab (path to IsaacLab), create an insert-drawer task in which a drawer (a URDF path), and a cube with a side length of 4.5 cm are placed on top of a table. A Franka robot (a URDF path) is positioned at the left-back corner of the table. The robot should pick up the cube placed in front of it, insert the cube into the opened drawer located near the center of the table, and then close the drawer after the cube has been placed inside. 
The task is considered successful if the cube remains inside the drawer and the drawer is largely closed.
\end{tcolorbox}

\begin{tcolorbox}[
    colback=blue!2,
    colframe=blue!20,
    boxrule=0.5pt,
    arc=2pt,
    left=6pt,
    right=6pt,
    top=5pt,
    bottom=5pt,
    fontupper=\ttfamily\small,
    title={\textnormal{\textsc{Stack-cube} (IsaacLab example)}},
    coltitle=black,
    colbacktitle=blue!6
]
/harbor:task-create In IsaacLab (path to IsaacLab), create a stack-cube task in which three cubes, each with a side length of 4.5 cm, are placed on top of a table as a triangle. A Franka robot (a URDF path) is positioned at the left-back corner of the table. The robot should first stack the closest cube on top of the middle cube, and then stack the remaining cube to the existing tower. The task is considered successful if all three cubes are stably stacked and there is no contact between the cubes and the robot.
\end{tcolorbox}

\begin{tcolorbox}[
    colback=blue!2,
    colframe=blue!20,
    boxrule=0.5pt,
    arc=2pt,
    left=6pt,
    right=6pt,
    top=5pt,
    bottom=5pt,
    fontupper=\ttfamily\small,
    title={\textnormal{\textsc{Lift-box} (IsaacLab example)}},
    coltitle=black,
    colbacktitle=blue!6
]
/harbor:task-create In IsaacLab (path to IsaacLab), create a lift-box task in which a box (path to box mesh) is placed on top of a table and at the center of the table. Two Franka robots (a URDF path) are positioned at the left-back and right-back corners of the table. The robots should reach the box sides, grasp it securely, and lift it 25 cm above the table surface. 
The task is considered successful if the box is lifted to the target height while stably grasped by the robot.
\end{tcolorbox}

\begin{tcolorbox}[
    colback=blue!2,
    colframe=blue!20,
    boxrule=0.5pt,
    arc=2pt,
    left=6pt,
    right=6pt,
    top=5pt,
    bottom=5pt,
    fontupper=\ttfamily\small,
    title={\textnormal{\textsc{Dex-grasp} (IsaacLab example)}},
    coltitle=black,
    colbacktitle=blue!6
]
/harbor:task-create In IsaacLab (path to IsaacLab), create a dexterous-grasp task in which an object is placed on top of a table in front of a robot with a dexterous hand (a URDF path). The robot should use its fingers to reach, make stable contact with the object, and lift it securely to a target position. The task is considered successful if the object reaches the target position within 10 cm distance threshold and remains stably held by the dexterous hand.
\end{tcolorbox}

\subsection{Performance for End-to-end Simulation Experiments}
\label{app:curves}

We provide the learning performance for the end-to-end simulation experiments summarized in Sec.~\ref{sec:result} and Tab.~\ref{tab:sim2real_reward_eval}. We evaluate HARBOR on four manipulation tasks, \texttt{Insert-Drawer}, \texttt{Stack-Cube}, \texttt{Lift-Box}, and \texttt{Dex-Grasp}, across four simulators: IsaacLab, ManiSkill, Genesis, and MJLab. For each task, HARBOR starts from the task specification and automates dependency setup, task implementation, reward generation, RL integration, and policy training. We plot both cumulative episodic return and binary success rate.
% to separate dense reward learning progress from task-level completion. 

\begin{figure}[t!]
    \centering
    \includegraphics[width=\textwidth]{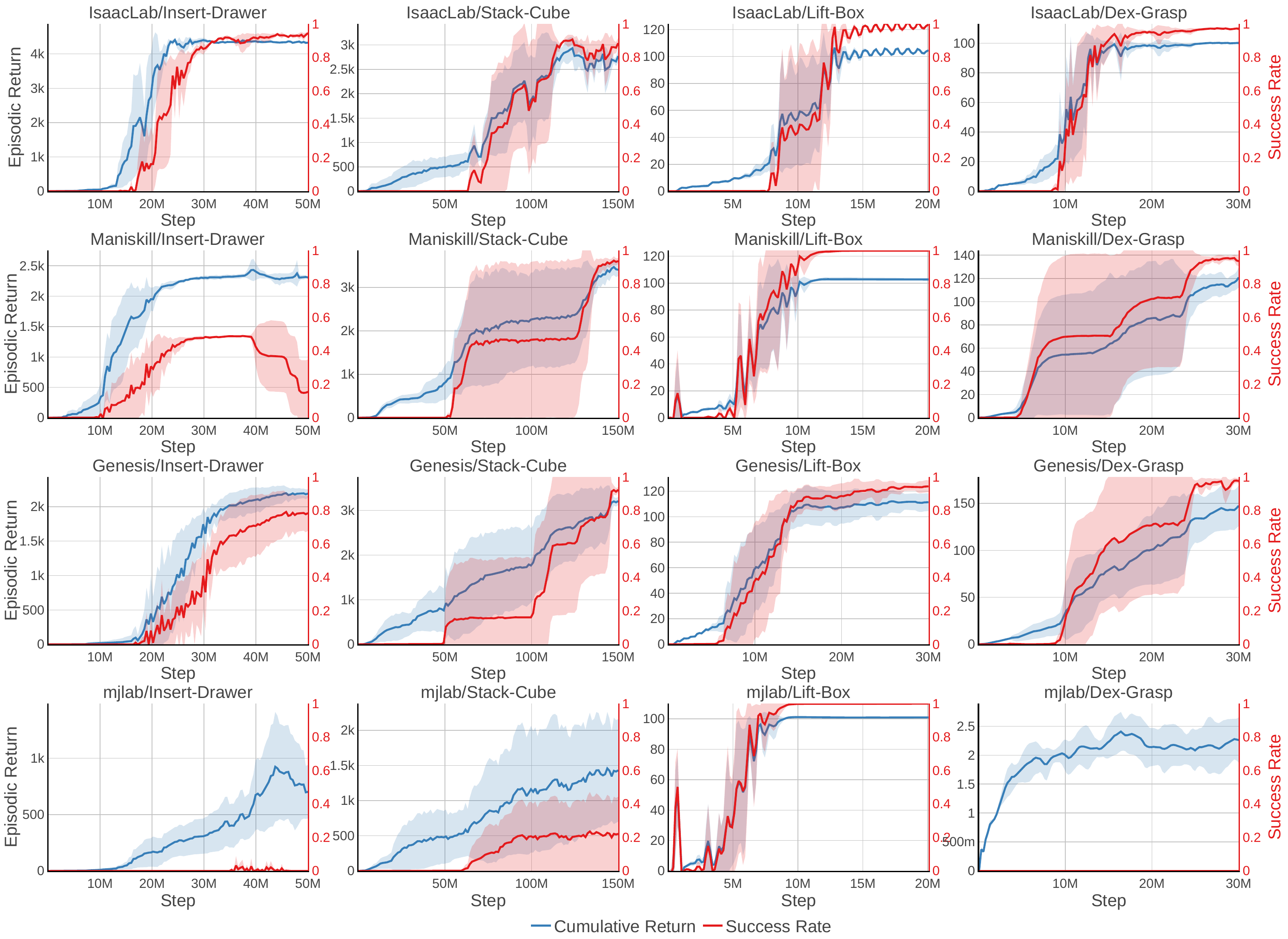}
    \vspace{-6mm}
    \caption{
    \textbf{End-to-end simulation learning curves across simulators and tasks.}
    We evaluate HARBOR-generated RL workflows on 16 simulator-task settings from IsaacLab, ManiSkill, Genesis, and MJLab. 
    Blue curves show cumulative episodic return and red curves show task success rate, with shaded regions indicating variation across seeds. 
    HARBOR generally produces policies with clear learning progress across simulators, and most tasks reach high success rates after training.
    }
    \label{fig:app-curve-benchmark}
    \includegraphics[width=\textwidth]{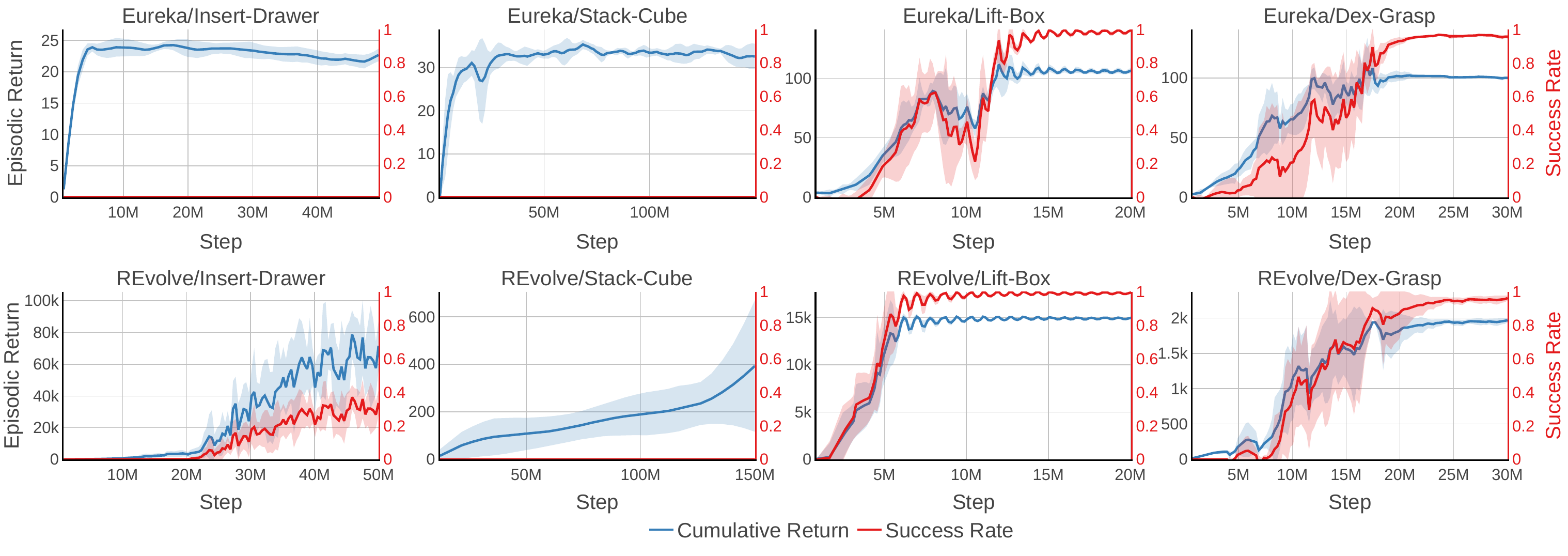}
    \vspace{-6mm}
    \caption{
    \textbf{Reward design baseline's learning curves on IsaacLab tasks.}
    Blue curves show cumulative episodic return and red curves show task success rate for Eureka and REvolve across four tasks. Both baselines perform well on Lift-Box and Dex-Grasp, while Insert-Drawer and Stack-Cube expose failure cases.
    }
    \label{fig:app-curve-baseline}
\end{figure}

Fig.~\ref{fig:app-curve-benchmark} shows the learning curves across all 16 simulator--task combinations. On IsaacLab, HARBOR achieves strong and stable learning across all four tasks: success rates converge rapidly for \texttt{Insert-Drawer}, \texttt{Lift-Box}, and \texttt{Dex-Grasp}, and the more long-horizon \texttt{Stack-Cube} task also reaches high success after staged reward optimization. ManiSkill and Genesis show similar trends: most tasks exhibit both increasing returns and high final success while also have larger variance according to the shadow area. In addition, \texttt{Lift-Box} and \texttt{Dex-Grasp} are consistently solved across these simulators, suggesting that HARBOR can effectively solve short-horizon tasks while adapting reward and implementation details to different simulator APIs and contact models.

We also provide baseline learning curves for Eureka and REvolve on the IsaacLab tasks in Fig.~\ref{fig:app-curve-baseline}. These curves explain the task-level differences in Tab.~\ref{tab:sim2real_reward_eval}. Both baselines perform well on \texttt{Lift-Box} and \texttt{Dex-Grasp}, where the reward structure is relatively direct. However, \texttt{Insert-Drawer} and \texttt{Stack-Cube} expose their limitations. Eureka fails to obtain nonzero success on both tasks, and REvolve only partially improves \texttt{Insert-Drawer} while failing on \texttt{Stack-Cube}. In contrast, HARBOR succeeds on both tasks by using rollout-behavior feedback to diagnose failure modes and by reusing human-provided heuristics to construct staged rewards for long-horizon behaviors. These results support the main-body conclusion that HARBOR's advantage is not only reward-code generation, but the surrounding harness that validates rollouts, preserves tuning history, and reuses experience across iterations.

\subsection{Detailed Results for RL-Tuning Experiments}
\label{app:rl_tuning_table}
We provide the full quantitative results for the RL-tuning experiments summarized in Sec.~\ref{sec:tuning}. As a recap, we evaluate HARBOR on three benchmark suites: IsaacLab~\citep{mittal2025isaac}, Loco-Mujoco~\citep{alhafez2023b}, and Bi-DexHands\citep{chen2022towards}. For each task, we compare the default configuration of PPO (from IsaacLab), SAC, and TD3 (from PQL~\citep{li2023parallel}) with the best HARBOR-tuned configuration selected from the tuning runs. All reported values are averaged over 5 seeds. We use three metrics to characterize tuning performance. 
\begin{itemize}[leftmargin=*]
    \item \textbf{AUC$_{0:N}$} denotes the area under the learning curve up to the training horizon $N$, and measures sample efficiency over the full training process (namely in practice we set $N$ equals the max step). For example, if two policies reach similar final returns but one obtains higher returns earlier in training, it will have a larger AUC$_{0:N}$, indicating better sample efficiency.
    \item \textbf{T$_\tau$} denotes the number of environment steps required to reach a task-specific return threshold $\tau$; for example, if $\tau$ is set to a return of 300 on a given task, a method that reaches 300 return at 5M steps is more sample-efficient than one that reaches it at 20M steps. Specifically, we define the return threshold \(\tau\) separately for each task and algorithm. For a given task-algorithm pair, we first apply the evaluation horizon used in the table and compute the final return of every Default and HARBOR run as the average return over the last 10\% of the truncated training curve. We then set
    \[
    \tau = 0.8 \cdot \max_i R^{\mathrm{final}}_i,
    \]where \(R^{\mathrm{final}}_i\) denotes the final return of run \(i\) among all Default and HARBOR seeds for that task-algorithm pair.
    \item \textbf{Final} denotes the mean episodic return over the last 10\% of training, and measures asymptotic performance. For example, if training runs for 50M steps, Final is computed from the returns observed in the last 5M steps, which indicates the converged performance after learning stabilizes.
\end{itemize}
Since reward scales differ across tasks and simulators, both AUC$_{0:N}$ and $\tau$ are computed separately for each task--algorithm pair. In the tables, ``--'' indicates that the corresponding configuration does not reach the threshold within the training horizon. Average T$_\tau$ improvement is computed only over task--algorithm pairs where both configurations reach $\tau$.

Tab.~\ref{tab:app-rl-tuning} shows that HARBOR consistently improves sample efficiency across simulators and algorithms. The improvements are most pronounced on IsaacLab and Loco-Mujoco, where several default configurations learn slowly or fail to reach the task-specific threshold, while HARBOR-tuned configurations achieve substantially stronger performance. On IsaacLab, HARBOR improves average AUC$_{0:N}$ by 1.3K\% and reaches thresholds 18.6\% faster on comparable pairs. On Loco-Mujoco, HARBOR improves average AUC$_{0:N}$ by 238\% and reduces T$_\tau$ by 60.4\%. On the more saturated Bi-DexHands benchmark, HARBOR still improves average AUC$_{0:N}$ by 12.3\% and reaches thresholds 42.2\% faster, while only falling short of SAC on \texttt{ShadowHandDoorCloseIn}. These results support the main-body observation that HARBOR improves early-stage learning and sample efficiency without requiring changes to the underlying RL algorithms.

\begin{table*}[t!]
\centering
\small
\setlength{\tabcolsep}{4.5pt}
\renewcommand{\arraystretch}{1.08}
\resizebox{0.8 \textwidth}{!}{
\begin{tabular}{llccccccc}
\toprule
\multirow{2}{*}{Task} &
\multirow{2}{*}{Alg.} &
\multicolumn{2}{c}{AUC$_{0:N}$ $\uparrow$} &
\multicolumn{2}{c}{T$_\tau$ $\downarrow$} &
\multicolumn{2}{c}{Final $\uparrow$} &
\multirow{2}{*}{Gain} \\
\cmidrule(lr){3-4}
\cmidrule(lr){5-6}
\cmidrule(lr){7-8}
& & Default & HARBOR & Default & HARBOR & Default & HARBOR & AUC \\
\midrule
\multirow{3}{*}{\texttt{AnymalC}}
& PPO & 13.3 & 49.5 & -- & 84.15M & 34.3 & 152 & +271.6\% \\
& SAC & 2.61 & 6.72 & -- & 46.60M & 3.52 & 9.16 & +157.7\% \\
& TD3 & 2.47 & 6.62 & -- & 34.30M & 3.59 & 8.12 & +168.1\% \\
\midrule
\multirow{3}{*}{\texttt{UnitreeA1}}
& PPO & 1.4K & 2.2K & 78.51M & 30.58M & 2.4K & 2.7K & +58.2\% \\
& SAC & 457 & 1.8K & -- & 50.89M & 609 & 2.2K & +302.6\% \\
& TD3 & 447 & 1.8K & -- & 51.51M & 654 & 2.4K & +301.1\% \\
\midrule
\multirow{3}{*}{\texttt{UnitreeGo2}}
& PPO & 1.8K & 2.5K & 53.45M & 21.50M & 2.8K & 2.9K & +37.1\% \\
& SAC & 386 & 2.4K & -- & 15.56M & 559 & 2.7K & +516.7\% \\
& TD3 & 398 & 2.3K & -- & 22.12M & 553 & 2.7K & +470.2\% \\
\midrule
\multirow{3}{*}{\texttt{UnitreeH1}}
& PPO & 183 & 450 & -- & 60.16M & 431 & 767 & +145.8\% \\
& SAC & 202 & 617 & -- & 63.22M & 333 & 911 & +205.7\% \\
& TD3 & 159 & 510 & -- & 54.32M & 257 & 828 & +220.0\% \\
\midrule
\multicolumn{2}{l}{\textbf{Average gain}} & \multicolumn{2}{c}{238\% AUC} & \multicolumn{2}{c}{60.4\% faster T$_\tau$} & \multicolumn{2}{c}{--} & -- \\
\bottomrule
\end{tabular}
}
\resizebox{\textwidth}{!}{
\begin{tabular}{llccccccc}
\toprule
\multirow{2}{*}{Task} &
\multirow{2}{*}{Alg.} &
\multicolumn{2}{c}{AUC$_{0:N}$ $\uparrow$} &
\multicolumn{2}{c}{T$_\tau$ $\downarrow$} &
\multicolumn{2}{c}{Final $\uparrow$} &
\multirow{2}{*}{Gain} \\
\cmidrule(lr){3-4}
\cmidrule(lr){5-6}
\cmidrule(lr){7-8}
& & Default & HARBOR & Default & HARBOR & Default & HARBOR & AUC \\
\midrule
\multirow{3}{*}{\texttt{Isaac-Factory-GearMesh-Direct-v0}}
& PPO & 133 & 435 & -- & 2.49M & 118 & 523 & +226.5\% \\
& SAC & 90.5 & 594 & -- & 2.45M & 109 & 716 & +556.7\% \\
& TD3 & 70.3 & 557 & -- & 2.86M & 81.6 & 696 & +693.2\% \\
\midrule
\multirow{3}{*}{\texttt{Isaac-Factory-PegInsert-Direct-v0}}
& PPO & 71.8 & 247 & -- & 2.32M & 87.4 & 280 & +244.0\% \\
& SAC & 48.5 & 277 & -- & 2.75M & 59.2 & 344 & +470.4\% \\
& TD3 & 44.9 & 319 & -- & 1.83M & 56.1 & 372 & +610.1\% \\
\midrule
\multirow{3}{*}{\texttt{Isaac-Franka-Cabinet-Direct-v0}}
& PPO & 2.1K & 2.3K & 24.16M & 19.66M & 2.7K & 2.8K & +8.4\% \\
& SAC & 1.4K & 2.0K & -- & 33.22M & 1.9K & 2.6K & +43.1\% \\
& TD3 & 1.3K & 2.3K & -- & 15.33M & 2.0K & 2.9K & +76.6\% \\
\midrule
\multirow{3}{*}{\texttt{Isaac-Lift-Cube-Franka-v0}}
& PPO & -1.36 & 98.6 & -- & 14.46M & -0.41 & 122 & +7331.4\% \\
& SAC & -1.07 & 47.4 & -- & 29.78M & -0.76 & 79.0 & +4535.7\% \\
& TD3 & -0.93 & 12.0 & -- & 19.46M & -0.35 & 20.6 & +1392.8\% \\
\midrule
\multicolumn{2}{l}{\textbf{Average gain}} & \multicolumn{2}{c}{1.3K\% AUC} & \multicolumn{2}{c}{18.6\% faster T$_\tau$} & \multicolumn{2}{c}{--} & -- \\
\bottomrule
\end{tabular}
}
\resizebox{\textwidth}{!}{
\begin{tabular}{llccccccc}
\toprule
\multirow{2}{*}{Task} &
\multirow{2}{*}{Alg.} &
\multicolumn{2}{c}{AUC$_{0:N}$ $\uparrow$} &
\multicolumn{2}{c}{T$_\tau$ $\downarrow$} &
\multicolumn{2}{c}{Final $\uparrow$} &
\multirow{2}{*}{Gain} \\
\cmidrule(lr){3-4}
\cmidrule(lr){5-6}
\cmidrule(lr){7-8}
& & Default & HARBOR & Default & HARBOR & Default & HARBOR & AUC \\
\midrule
\multirow{3}{*}{\texttt{ShadowHandBlockStack}}
& PPO & 493 & 555 & 65.89M & 44.26M & 716 & 656 & +12.5\% \\
& SAC & 165 & 190 & 19.75M & 4.77M & 189 & 193 & +14.7\% \\
& TD3 & 190 & 199 & 11.10M & 3.33M & 209 & 181 & +4.7\% \\
\midrule
\multirow{3}{*}{\texttt{ShadowHandDoorCloseIn}}
& PPO & 287 & 295 & 5.20M & 3.56M & 293 & 312 & +3.1\% \\
& SAC & 240 & 236 & 8.69M & 13.41M & 257 & 267 & -1.7\% \\
& TD3 & 237 & 281 & -- & 14.61M & 250 & 296 & +18.2\% \\
\midrule
\multirow{3}{*}{\texttt{ShadowHandDoorOpenIn}}
& PPO & 395 & 399 & -- & 34.71M & 412 & 487 & +0.9\% \\
& SAC & 283 & 319 & 15.87M & 7.44M & 345 & 359 & +12.5\% \\
& TD3 & 319 & 348 & 6.94M & 2.83M & 354 & 368 & +9.0\% \\
\midrule
\multirow{3}{*}{\texttt{ShadowHandOver}}
& PPO & 17.7 & 26.1 & 34.08M & 9.39M & 26.8 & 30.1 & +47.8\% \\
& SAC & 19.2 & 22.3 & 38.14M & 15.07M & 23.1 & 27.9 & +16.5\% \\
& TD3 & 17.9 & 19.6 & 15.19M & 12.08M & 23.3 & 24.6 & +9.5\% \\
\midrule
\multicolumn{2}{l}{\textbf{Average gain}} & \multicolumn{2}{c}{12.3\% AUC} & \multicolumn{2}{c}{42.2\% faster T$_\tau$} & \multicolumn{2}{c}{--} & -- \\
\bottomrule
\end{tabular}
}
\caption{
\textbf{Detailed RL-tuning results across Loco-MuJoCo, IsaacLab, and Bi-DexHands tasks.}
For each task and algorithm, we compare the default configuration with the best HARBOR-tuned configuration over 5 seeds.
% AUC$_{0:N}$ measures sample efficiency over the full training horizon, T$_\tau$ reports the number of environment steps required to reach a task-specific performance threshold, and Final reports the mean return over the last 10\% of training.
% Higher AUC$_{0:N}$ and Final are better; lower T$_\tau$ is better.
% ``--'' indicates that the threshold is not reached within the training horizon.
}
\label{tab:app-rl-tuning}
\end{table*}

\subsection{Ablation Setup and Profiling Details}
\label{app:ablation}

This section expands the pillar-ablation study of Sec.~\ref{sec:ablation} (Tab.~\ref{tab:workflow_ablation_transposed_clean}).

\paragraph{Stages and configurations.} The pipeline is reported over five stages---dependency and benchmark setup (jointly), then RL integration, task generation, reward generation, and RL tuning---under five configurations: the full harness (\textbf{\textsc{Harbor}}) and three single-pillar ablations of Sec.~\ref{sec:HARBOR}, namely \emph{w/o CCDE} (removing centralized-control/decentralized-execution, i.e., no subagent isolation or parallel trial execution), \emph{w/o gate} (no verifiable smoke gates), and \emph{w/o experience} (authored from scratch, without templates, references, or heuristics), plus a no-harness \emph{Vanilla} baseline (a single-context agent with none of the three pillars).

\paragraph{Metrics and profiling.} For each stage we report S.R.\ (successes over 10 repeated runs), wall-clock time (minutes), the context tax Ctx (cache-read tokens re-read on every turn, in M---our proxy for the cost of carrying accumulated context), and estimated cost (USD); Time, Ctx, and cost are summed over the five stages and S.R.\ over the 50 runs. Each stage is profiled as an independent \texttt{claude-code-profiler} window, and we compare on the cross-group--comparable $\Sigma$ stage-windows figures (the five per-stage windows summed, excluding idle and orchestration). Costs are estimates under mixed-model pricing on a single ManiSkill slice and should be read as ratios, not absolute levels. Lacking subagent parallelism, \emph{w/o CCDE} and \emph{Vanilla} run the reward and RL-tuning trials serially, so their time, cache-read, and cost on those two stages are scaled $\times4$.

\paragraph{Mechanisms behind the three findings.} For \emph{w/o CCDE}, the single window's context tax grows super-linearly with pipeline length, and serial trials make it second-slowest overall behind only \emph{Vanilla}; the over-long context is what corrupts its late-stage edits. The \emph{w/o gate} render-path defect ($^{\dagger}$ in Table~\ref{tab:workflow_ablation_transposed_clean}) produces zero rendered frames while the process exits $0$---invisible to static checks, caught only by a render smoke. \emph{w/o experience} must re-derive empirical contracts from scratch, most notably the ManiSkill auto-reset/truncation convention. The full cost ordering is \textsc{Harbor} ($\sim$\$71) $<$ \emph{w/o gate} ($\sim$\$80) $<$ \emph{w/o experience} ($\sim$\$120) $<$ \emph{w/o CCDE} ($\sim$\$131) $<$ \emph{Vanilla} ($\sim$\$182): the full harness is the cheapest measured configuration despite running every gate, consistent with the Pareto-optimality claim in Sec.~\ref{sec:ablation}.

\subsection{Sonnet Ablation}
\label{app:sonnet}

To probe whether HARBOR's benefits depend on the strongest available model, we consider repeating the pillar-ablation study of Sec.~\ref{sec:ablation} on the ManiSkill \texttt{Push-Cube} pipeline with Sonnet as the underlying agent model, keeping the same five stages, five configurations, default $N=4$ parallelization, and per-stage profiling protocol of Appx.~\ref{app:ablation}. We conclude four mechanisms implied by the main experiments:

\begin{itemize}[leftmargin=*]
\item \textbf{Reliability drops, concentrated in reasoning-heavy stages.} A weaker base model lowers per-stage success most where the task is generative and long-horizon---RL integration, reward generation, and RL tuning---while the mechanical dependency-setup stage stays near-saturated. We therefore lower S.R. mainly on those three stages and leave dependency setup almost unchanged.
\item \textbf{The harness's relative value grows.} Because isolation, gates, and experience compensate for a weaker model, the \textsc{Harbor}$-$\emph{Vanilla} reliability gap widens from $19$ points under Opus ($48$ vs.\ $29$) to $\sim\!30$ points under Sonnet ($48$ vs.\ $18$): the full harness degrades gracefully while the no-harness baseline collapses on the hard stages.
\item \textbf{Cost falls by ${\sim}3\times$ but the ordering is preserved.} Sonnet's ${\sim}5\times$ lower per-token price is partly offset by ${\sim}1.5\times$ more retries on the hard stages, so each configuration's cost scales by roughly $0.32$; the cheap-to-expensive ordering (\textsc{Harbor} $<$ w/o gate $<$ w/o experience $<$ w/o CCDE $<$ Vanilla) is unchanged.
\item \textbf{Time and context tax rise modestly.} More retries inflate wall-clock time (${\sim}1.15\times$) and the context tax (${\sim}1.3\times$), but subagent isolation still caps context growth, so \emph{w/o CCDE} and \emph{Vanilla} remain worst on Ctx and time. The render-path defect ($^{\dagger}$) is a tooling/gate gap independent of model capability, so \emph{w/o gate} still passes it silently.
\end{itemize}

\begin{table*}[t]
\centering
\footnotesize
\setlength{\tabcolsep}{3.6pt}
\renewcommand{\arraystretch}{1.18}
\caption{Ablations on the ManiSkill \texttt{Push-Cube} pipeline. $^{\dagger}$ marks a stage that reported success but hid a silent render-path defect. We use default $N=4$ for parallelization and Sonnet 4.6}
\label{tab:sonnet_ablation}
\resizebox{\textwidth}{!}{%
\begin{tabular}{
>{\centering\arraybackslash}p{2.7cm}
ccc
ccc
ccc
ccc
ccc
}
\toprule
\textbf{Stage}
& \multicolumn{3}{c}{Vanilla}
& \multicolumn{3}{c}{\textsc{Harbor}}
& \multicolumn{3}{c}{w/o CCDE}
& \multicolumn{3}{c}{w/o gate}
& \multicolumn{3}{c}{w/o experience} \\
\cmidrule(lr){2-4}
\cmidrule(lr){5-7}
\cmidrule(lr){8-10}
\cmidrule(lr){11-13}
\cmidrule(lr){14-16}
& S.R. & Time & Ctx/\$
& S.R. & Time & Ctx/\$
& S.R. & Time & Ctx/\$
& S.R. & Time & Ctx/\$
& S.R. & Time & Ctx/\$ \\
\midrule
Dependency setup
& 9/10 & 25.1 & 6.21M / \$5.73
& 10/10 & 12.9 & 3.61M / \$2.62
& 10/10 & 14.3 & 5.67M / \$4.91
& 10/10 & 15.1 & 2.73M / \$4.08
& 8/10 & 12.2 & 0.88M / \$1.13 \\

RL integration
& 1/10 & 34.2 & 19.71M / \$10.05
& 10/10 & 19.7 & 4.84M / \$2.92
& 8/10 & 20.2 & 15.73M / \$7.88
& 9/10$^{\dagger}$ & 4.7 & 1.68M / \$1.13
& 5/10 & 17.0 & 9.18M / \$4.42 \\

Task generation
& 5/10 & 15.9 & 6.86M / \$4.09
& 10/10 & 9.1 & 3.80M / \$2.68
& 8/10 & 4.5 & 5.19M / \$2.59
& 7/10 & 6.0 & 1.47M / \$1.29
& 5/10 & 12.1 & 2.47M / \$2.09 \\

Reward generation
& 1/10 & 56.6 & 33.35M / \$15.39
& 9/10 & 4.8 & 18.66M / \$9.92
& 6/10 & 27.6 & 32.24M / \$14.22
& 3/10 & 3.3 & 9.17M / \$5.73
& 1/10 & 6.9 & 13.33M / \$9.30 \\

RL tuning
& 2/10 & 81.7 & 41.47M / \$23.02
& 9/10 & 4.3 & 5.59M / \$4.54
& 6/10 & 29.3 & 26.07M / \$12.21
& 3/10 & 8.7 & 16.12M / \$13.28
& 2/10 & 13.3 & 31.85M / \$21.55 \\

\midrule
\textbf{Total ($\Sigma$-win.)}
& 18/50 & 213.5 & 107.60M / \$58.28
& 48/50 & 50.8 & 36.50M / \$22.68
& 38/50 & 95.9 & 84.90M / \$41.81
& 32/50$^{\dagger}$ & 37.8 & 31.17M / \$25.51
& 21/50 & 61.5 & 57.71M / \$38.49 \\
\bottomrule
\end{tabular}
}
\end{table*}

\end{document}